\RequirePackage{fix-cm}
\documentclass[twocolumn]{svjour3}          
\smartqed  
\usepackage{graphicx}
\usepackage{mathptmx}
\usepackage{xcolor}
\usepackage{paralist}
\usepackage[sort&compress,numbers,square]{natbib}
\usepackage{url}
\DeclareGraphicsExtensions{.jpg,.png}
\newcommand{\eg}{\emph{e.g.},}

\newcommand{\etal}{\emph{et al.}}

\journalname{Arxiv}

\begin{document}

\title{Collecting and Annotating the Large Continuous Action Dataset}

\titlerunning{The Large Continuous Action Dataset}

\author{Daniel Paul Barrett \and
        Ran Xu \and
        Haonan Yu \and
        Jeffrey Mark Siskind}

\authorrunning{Daniel Paul Barrett \etal}

\institute{Daniel Paul Barrett \at
              School of Electrical and Computer Engineering, Purdue University
           \and
           Ran Xu \at
              Computer Science and Engineering, SUNY Buffalo
           \and
           Haonan Yu \at
              School of Electrical and Computer Engineering, Purdue University
           \and
           Jeffrey Mark Siskind \at
              School of Electrical and Computer Engineering, Purdue University\\
              Tel.: +1-765-496-3197\\
              Fax: +1-765-494-2706\\
              \email{qobi@purdue.edu}}

\date{Received: date / Accepted: date}

\maketitle

\begin{abstract}
We make available to the community a new dataset to support action-recognition
research.
This dataset is different from prior datasets in several key ways.
It is significantly larger.
It contains streaming video with long segments containing multiple action
occurrences that often overlap in space and/or time.
All actions were filmed in the same collection of backgrounds so that
background gives little clue as to action class.
We had five humans replicate the annotation of temporal extent of action
occurrences labeled with their class and measured a surprisingly low level of
intercoder agreement.
A baseline experiment shows that recent state-of-the-art methods perform poorly
on this dataset.
This suggests that this will be a challenging dataset to foster advances in
action-recognition research.
This manuscript serves to describe the novel content and characteristics of the
LCA dataset, present the design decisions made when filming the dataset, and
document the novel methods employed to annotate the dataset.
\end{abstract}

\section{Introduction}
\label{sec:intro}

There has been considerable research interest in action recognition in video
over the past two decades \citep{YamotoOI92, zhu2013actons, Wang2009b,
  MajiActionCVPR11, Wang2013, dtijcv2013, wang:2011:inria-00583818:1,
  blank2005actions, gaidon2014hierarchies, Li2007, Brand1997, niebles2010,
  tian2013, Wang2009a, everts2013, liu2011attributes, Sadanand2012,
  willems2008efficient, ikizler2010, Liu2009, Kojima2002, gopalan2013sparsity,
  iccv2009MessingPalKautz, rodriguez2008action, schuldt2004recognizing,
  Gupta2007, Das2013, jain2013patches, SiskindM96, Hanckmann2012,
  yuan2013sparse, khan2012, tena2007, Yuan2013, Khan2011a,
  reddy2013recognizing, oneata2013fisher, Rohrbach2013, Moore1999, Yu2013,
  Guadarrama2013, oreifej2014trajectory, dominey2005, Khan2011b, song2013,
  ke2007, laptev2005space, tang2012, Siskind2000, Xu2002, Kuehne11,
  wu2011context, Barbu2012b, wang2013pose, Barbu2012a, Krishnamoorthy2013,
  siddharth2014}.
To support such research, numerous video datasets have been gathered.
Liu \etal\ \cite{liu2011} summarize the available datasets as of 2011.
%
%
%
These include KTH (6 classes, \citep{schuldt2004recognizing}), Weizmann (10
classes, \citep{blank2005actions}), CMU Soccer (7 classes, \citep{Efros2003}),
CMU Crowded (5 classes, \citep{ke2007}), UCF Sports (9 classes,
\citep{rodriguez2008action}), UR ADL (10 classes,
\citep{iccv2009MessingPalKautz}), UM Gesture (14 classes, \citep{Lin2009}), UCF
Youtube (11 classes, \citep{Liu2009}), Hollywood-1 (8 classes,
\citep{Laptev2008}), Hollywood-2 (12 classes, \citep{marszalek09}), MultiKTH (6
classes, \citep{Uemura2008}), MSR (3 classes, \citep{Yuan2009}), and TRECVID (10
classes, \citep{Smeaton2006}).
These datasets contain short clips, each depicting one of a small number of
classes (3--14).
Several more recent datasets also contain short clips, each depicting a single
action, but with a larger number of action classes: UCF50 (50 classes,
\citep{reddy2013recognizing}), HMDB51 (51 classes, \citep{Kuehne11}), and UCF101
(101 classes, \citep{soomro2012ucf101}).
The VIRAT dataset \citep{Oh2011} has 12~classes and longer streaming video.

Here, we introduce a new dataset called the \emph{Large Continuous Action
  Dataset} (LCA).\@
This dataset contains depictions of 24 action classes.
The video for this dataset was filmed and annotated as part of the DARPA
Mind's Eye program.
A novel characteristic of this dataset is that rather than consisting of short
clips each of which depicts a single action class, this dataset contains much
longer streaming video segments that each contain numerous instances of a
variety of action classes that often overlap in time and may occur in different
portions of the field of view.
The annotation that accompanies this dataset delineates not only which actions
occur but also their temporal extent.

Many of the prior datasets were culled from video downloaded from the internet.
In contrast, the LCA dataset contains video that was filmed specifically to
construct the dataset.
While the video was filmed with people hired to act out the specified actions
according to a general script, the fact that the video contains long streaming
segments tends to mitigate any artificial aspects of the video and render the
action depictions to be quite natural.
Moreover, the fact that all of the video was filmed in a relatively small
number of distinct backgrounds makes the dataset challenging; the background
gives little clue as to the action class.

A further distinguishing characteristic of the LCA dataset is the degree of
ambiguity.
Most prior action-recognition datasets, in fact most prior datasets for all
computer-vision tasks, make a tacit assumption that the labeling is unambiguous
and thus there is a `ground truth.'
We had a team of five human annotators each annotate the entire LCA dataset.
This allowed us to measure the degree of intercoder agreement.
Surprising, there is a significant level of disagreement between humans as to
the temporal extent of most action instances.
We believe that such inherent ambiguity is a more accurate reflection of the
underlying action-recognition task and hope that the multiplicity of divergent
annotations will help spur novel research with this more realistic dataset.

Another distinguishing characteristic of the LCA dataset is that some action
occurrences were filmed simultaneously with multiple cameras with partially
overlapping fields of view.
While the cameras were neither spatially calibrated nor temporally
synchronized, the fact that we have multiple annotations of the temporal extent
of action occurrences may support future efforts to perform temporal
synchronization after the fact.
Furthermore, while most of the video was filmed from ground-level cameras with
horizontal view, some of the video was filmed with aerial cameras with bird's
eye view.
Some of this video was filmed simultaneously with ground cameras.
This may support future efforts to conduct scene reconstruction.

Some datasets are provided with specific tasks and evaluation metrics.
We refrain from doing so for this dataset.
\emph{Inter alia}, we do not provide official sanctioned splits into validation
sets.
Instead, we leave it up to the community to make use of this dataset in a
creative fashion for as many different tasks as it will be suited.

In particular, the evaluations conducted by Mind's Eye included a specific set
of tasks, namely Recognition (REC), Description (DES), Gap Filling (GAP), and
Anomaly Detection (ANM) with specific evaluation metrics.
Such tasks and metrics are expressly \emph{not} part of LCA.\@
The evaluations conducted under Mind's Eye make use of material that is not
included in LCA and metrics that are not public.
Likewise, LCA contains material that was not available for use during the
Mind's Eye evaluations.
Thus said evaluations could not be replicated outside of the context of Mind's
Eye.
Any potential future evaluations conducted with LCA would thus be incomparable
to the results obtained under Mind's Eye.

The entire LCA dataset, including the video and the annotations, has been
cleared for release by DARPA.\@
The remaining material gathered by DARPA for the Mind's Eye Year~2 evaluation
that is not included in LCA may not have been cleared for release.
As part of the release process, some video was edited to remove certain
portions.
Furthermore, the annotation process was performed with the particular versions
of the videos included in LCA as provided by DARPA.\@
These may have been transcoded from the original as filmed by the camera.
Thus, the time alignment of the annotations can only be guaranteed with the
versions of the videos included in LCA.\@
The time alignment may not be correct for any other versions of these videos
that may be residual from the DARPA Mind's Eye program.

\section{Collection}
\label{sec:collection}

The video for this dataset was filmed by DARPA in conjunction with Mitre and
several performers from the Mind's Eye
program.\footnote{\url{http://www.visint.org/}}
The bulk of the video was filmed as part of the Mind's Eye Year~2 evaluation.
Within the Mind's Eye program, that video was referred to as C-D2a, C-D2b,
C-D2c, and the Y2 Evaluation dataset.
This video is disjoint from that gathered by Janus Research Group as part of
the Mind's Eye Year~1 evaluation.
Within the Mind's Eye program, that video was refereed to as C-D1a, C-D1b, C-D1,
and C-E1.
As the LCA dataset contains only a subset of that material, we refrain from
using all such terminology in reference to LCA.\@
Also note that the LCA dataset contains some video that was not included in
the data used as part of the Mind's Eye evaluations.

The LCA dataset was filmed at three different locations over four periods:
\begin{compactenum}
\item The Great Plains Joint Training Center (GPJTC), an army training facility
  in Kansas, on 22 August 2011.
  Filming took place in two contexts at GPJTC, a simulated country road and a
  simulated safe house.
\item Strategic Operations, Inc. (STOPS), a training facility in San Diego, on
  14--15 December 2011 and on 6--9 March 2012.
  %
  %
  Filming during the first period took place in three contexts: two different
  simulated country roads and one simulated safe house.
  Filming during the second period took place in five contexts: two different
  simulated country roads, two different simulated safe houses, and one
  other.
  %
  %
\item Fort Indiantown Gap (FITG), an army training facility in Pennsylvania, on
  19--20 June 2012.
\end{compactenum}

\begin{figure*}
  \centering
  \begin{tabular}{@{}c@{\hspace{2pt}}c@{\hspace{2pt}}c@{\hspace{2pt}}c@{\hspace{2pt}}c@{\hspace{2pt}}c@{}}
    \includegraphics[width=0.15\textwidth,natwidth=640,natheight=480]{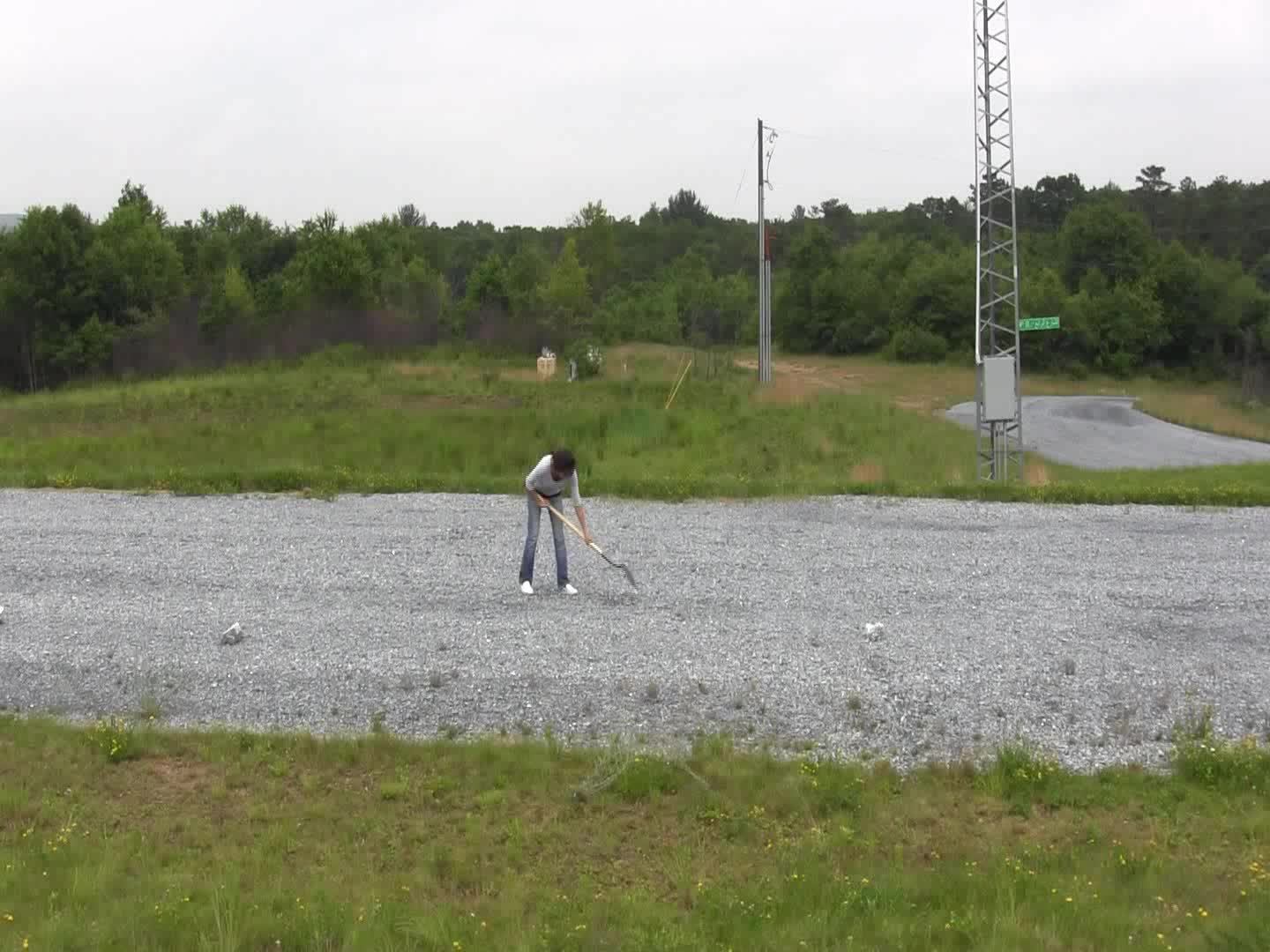}&
    \includegraphics[width=0.15\textwidth,natwidth=640,natheight=480]{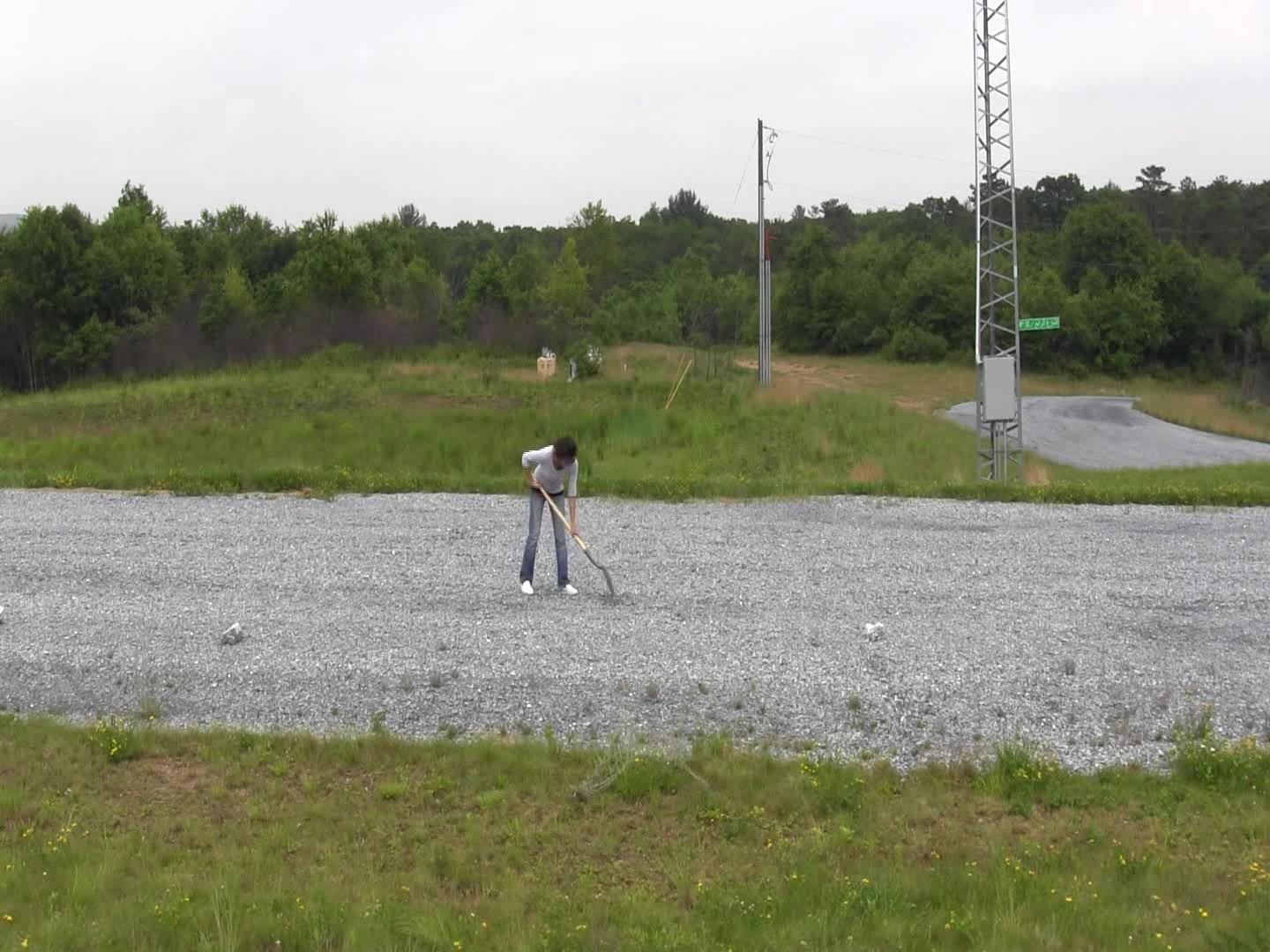}&
    \includegraphics[width=0.15\textwidth,natwidth=640,natheight=480]{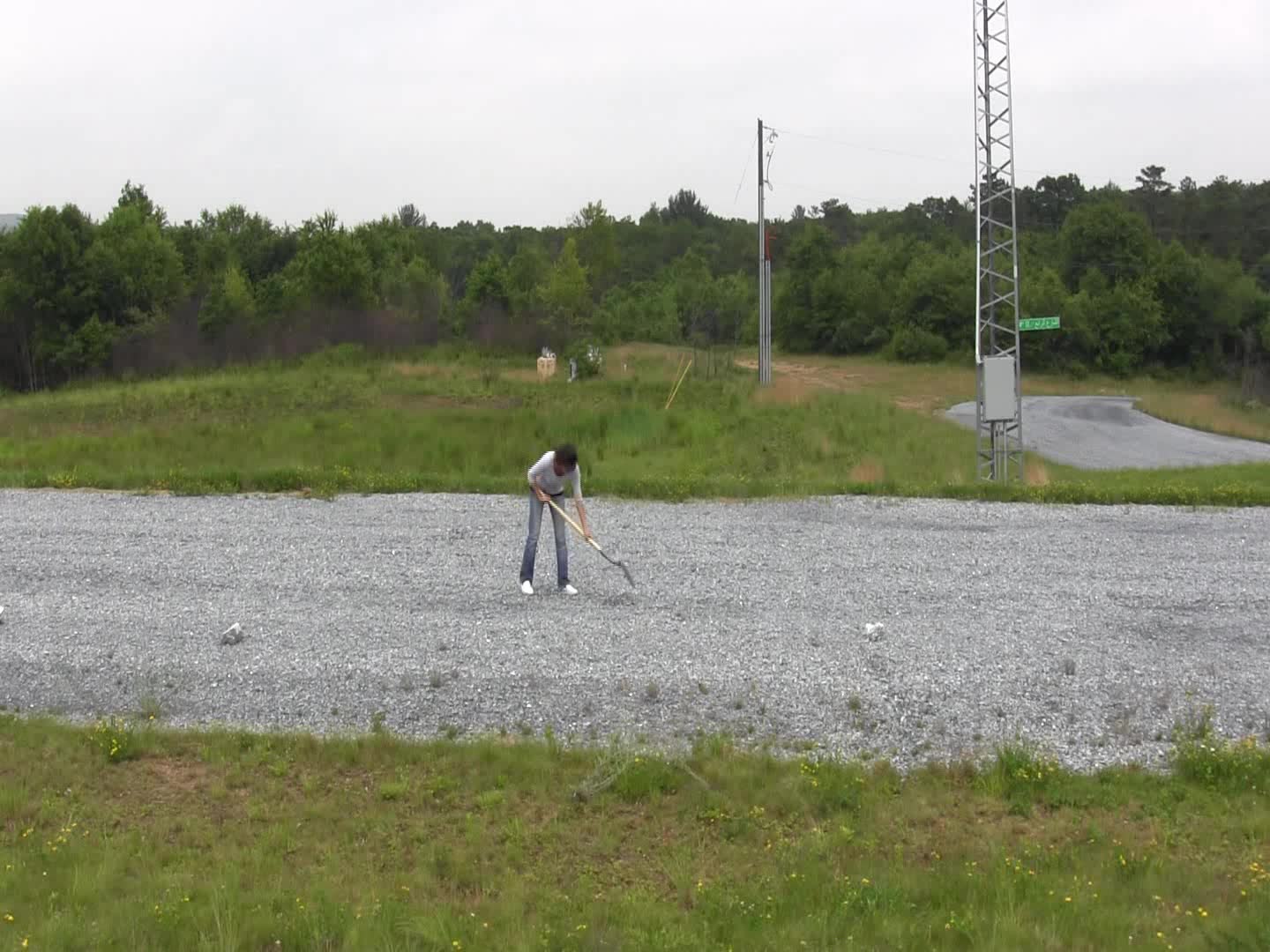}&
    \includegraphics[width=0.15\textwidth,natwidth=640,natheight=480]{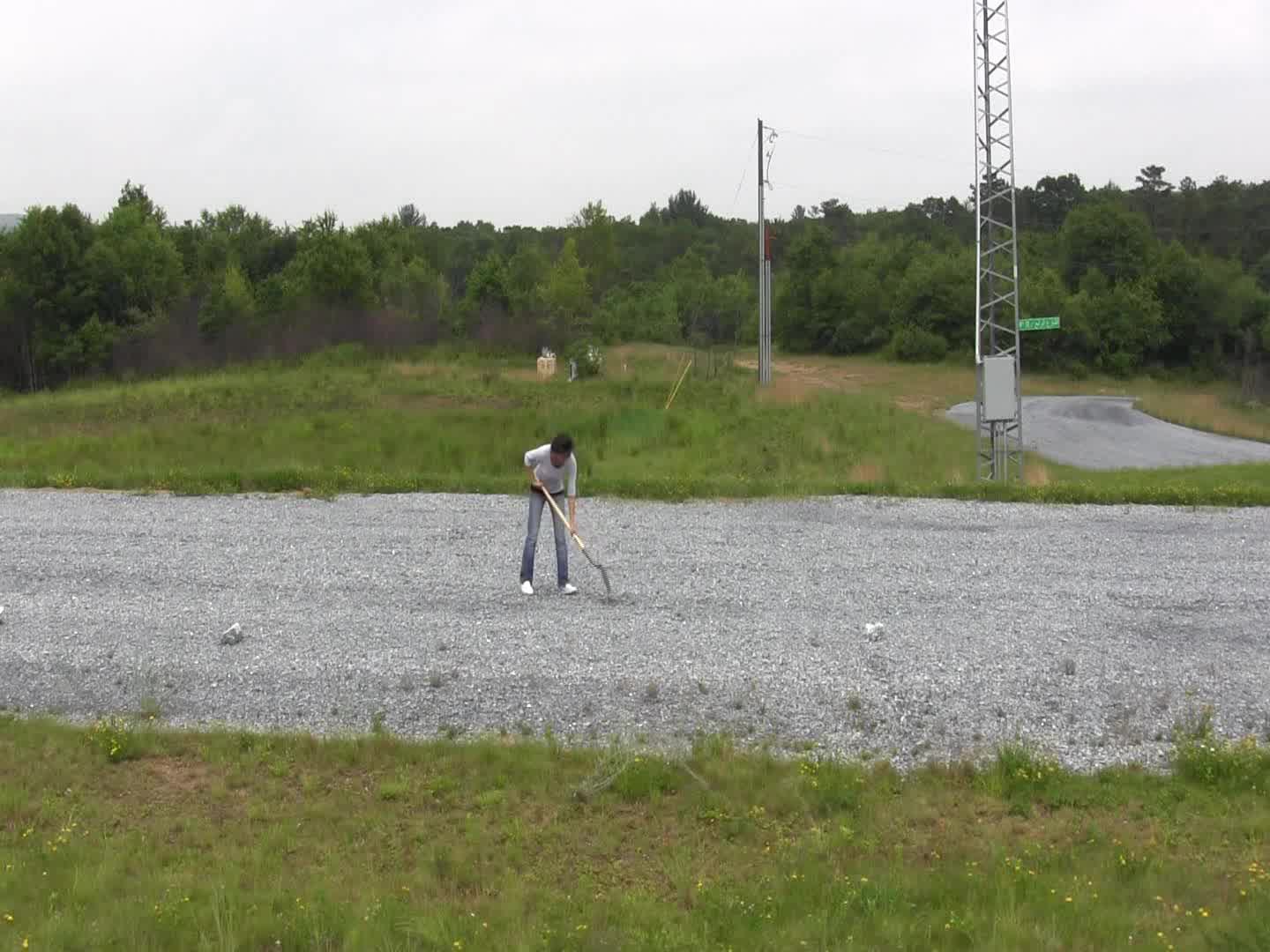}&
    \includegraphics[width=0.15\textwidth,natwidth=640,natheight=480]{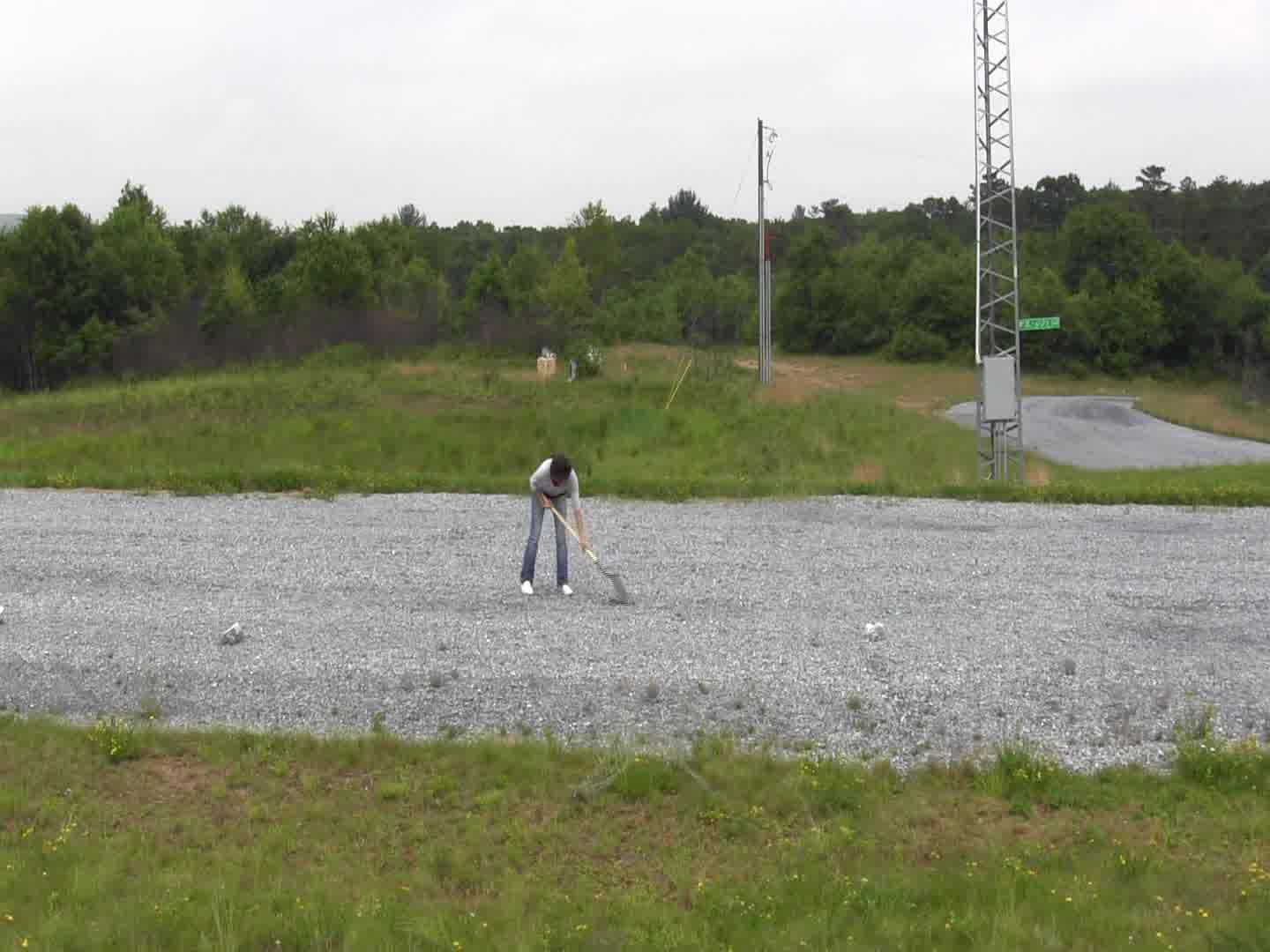}&
    \includegraphics[width=0.15\textwidth,natwidth=640,natheight=480]{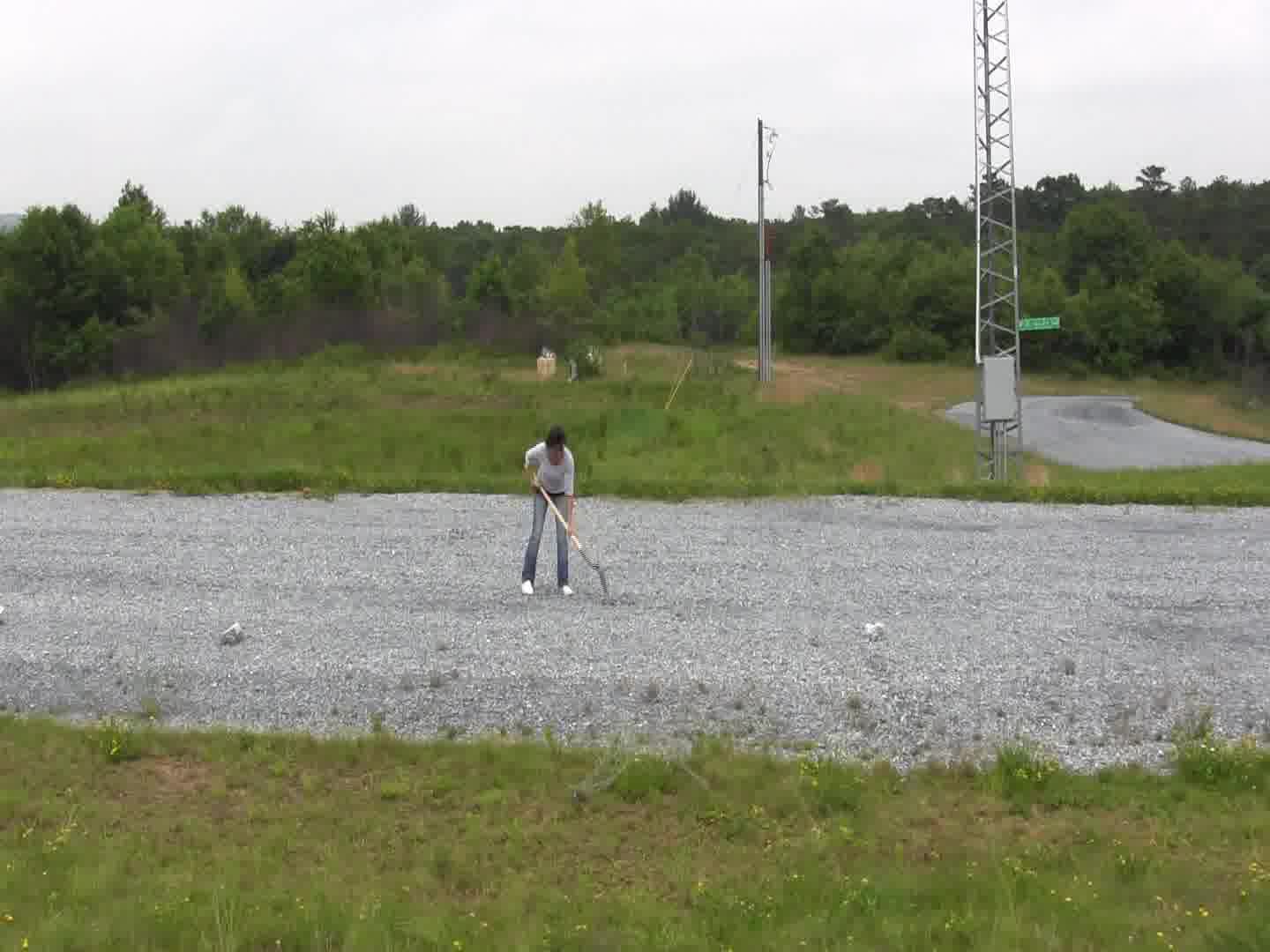}\\
    \includegraphics[width=0.15\textwidth,natwidth=640,natheight=480]{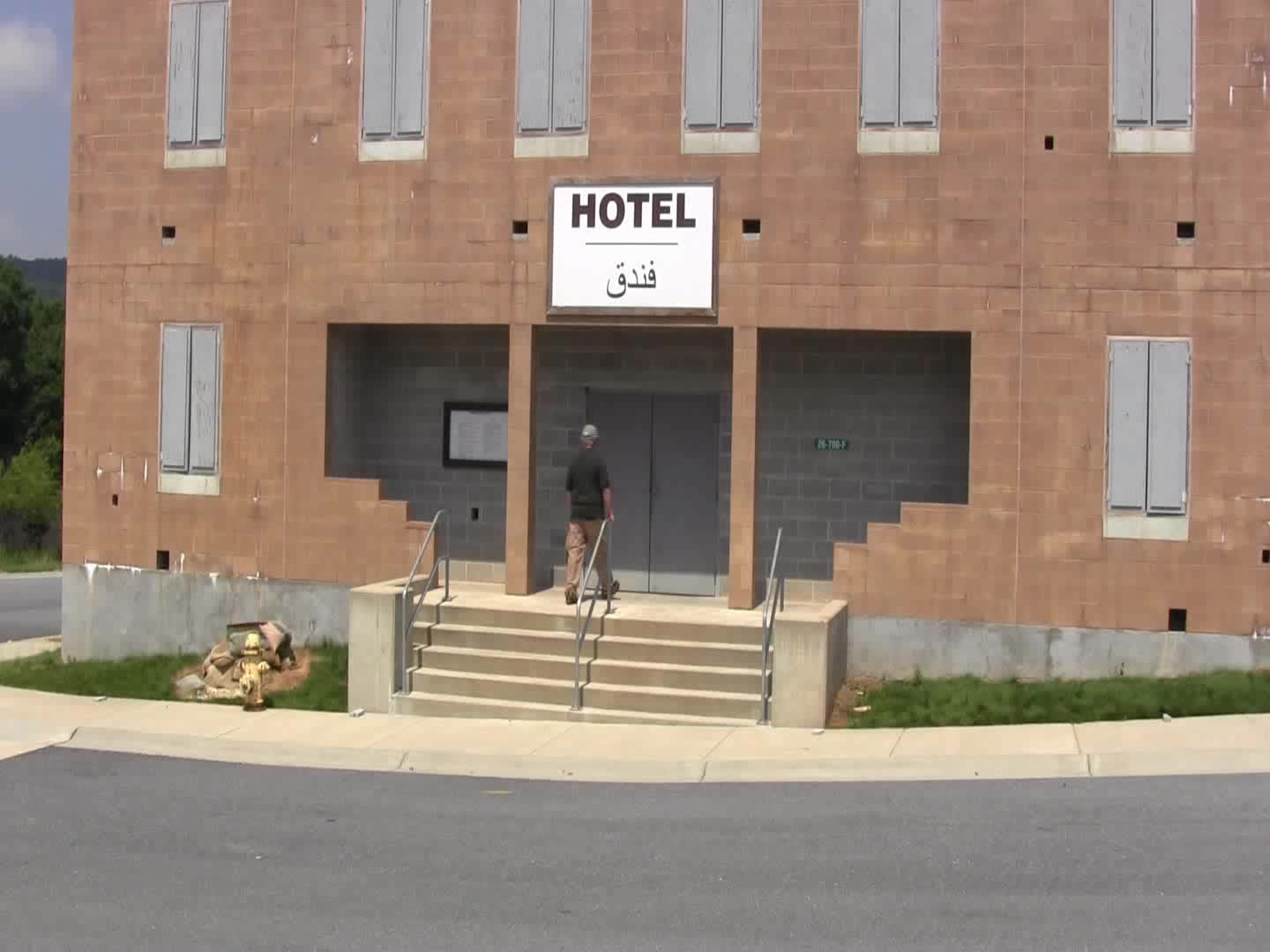}&
    \includegraphics[width=0.15\textwidth,natwidth=640,natheight=480]{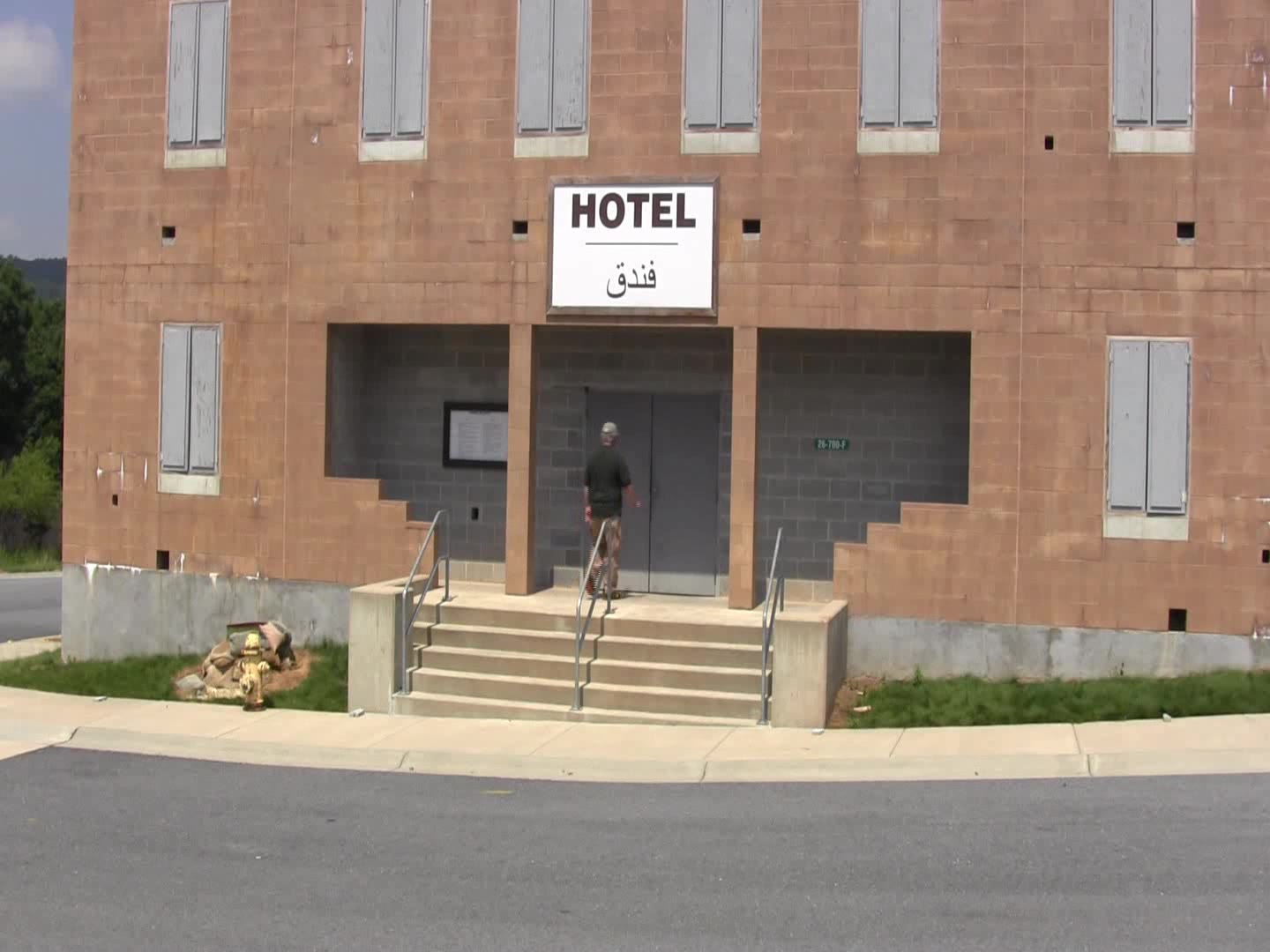}&
    \includegraphics[width=0.15\textwidth,natwidth=640,natheight=480]{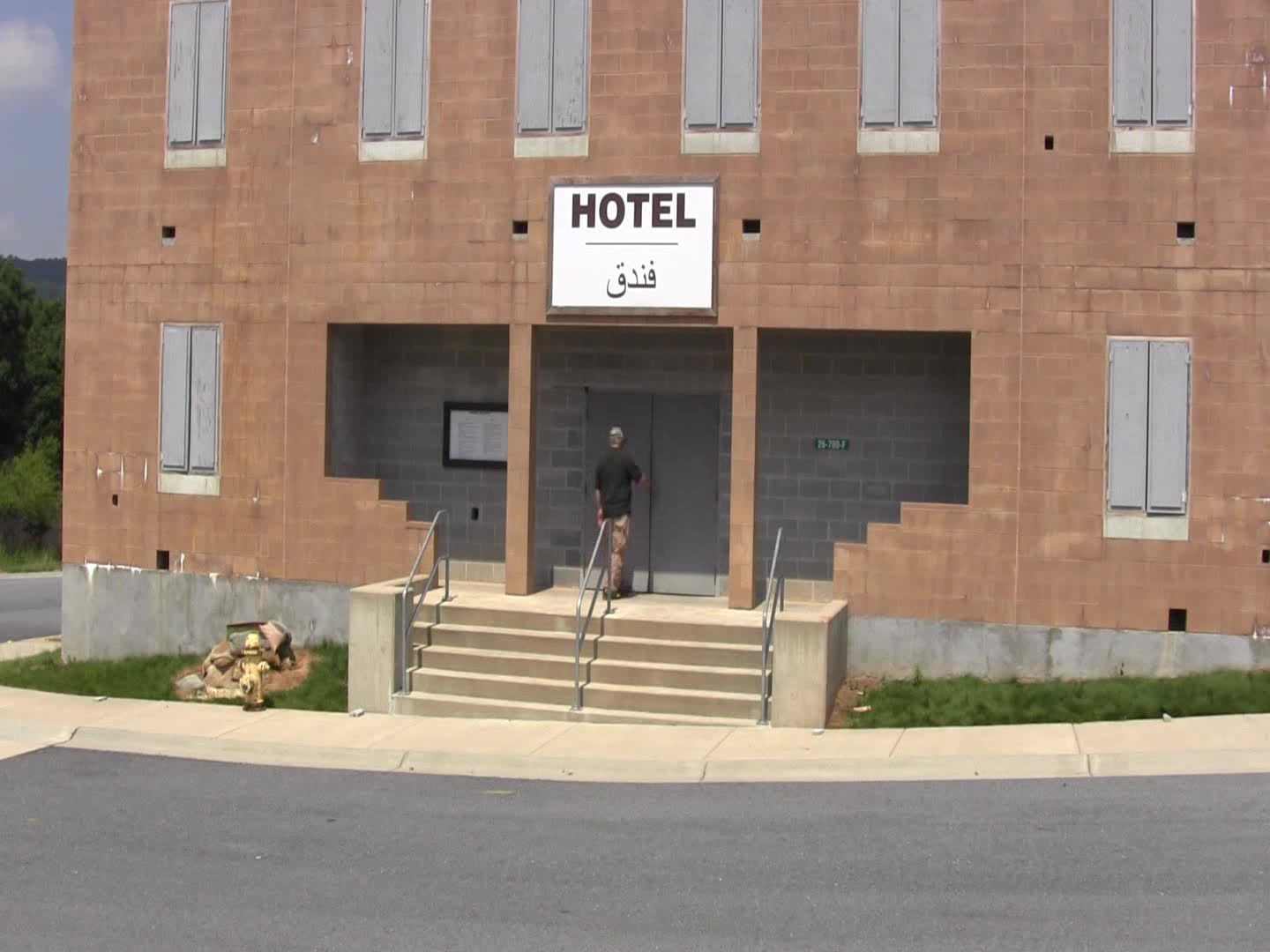}&
    \includegraphics[width=0.15\textwidth,natwidth=640,natheight=480]{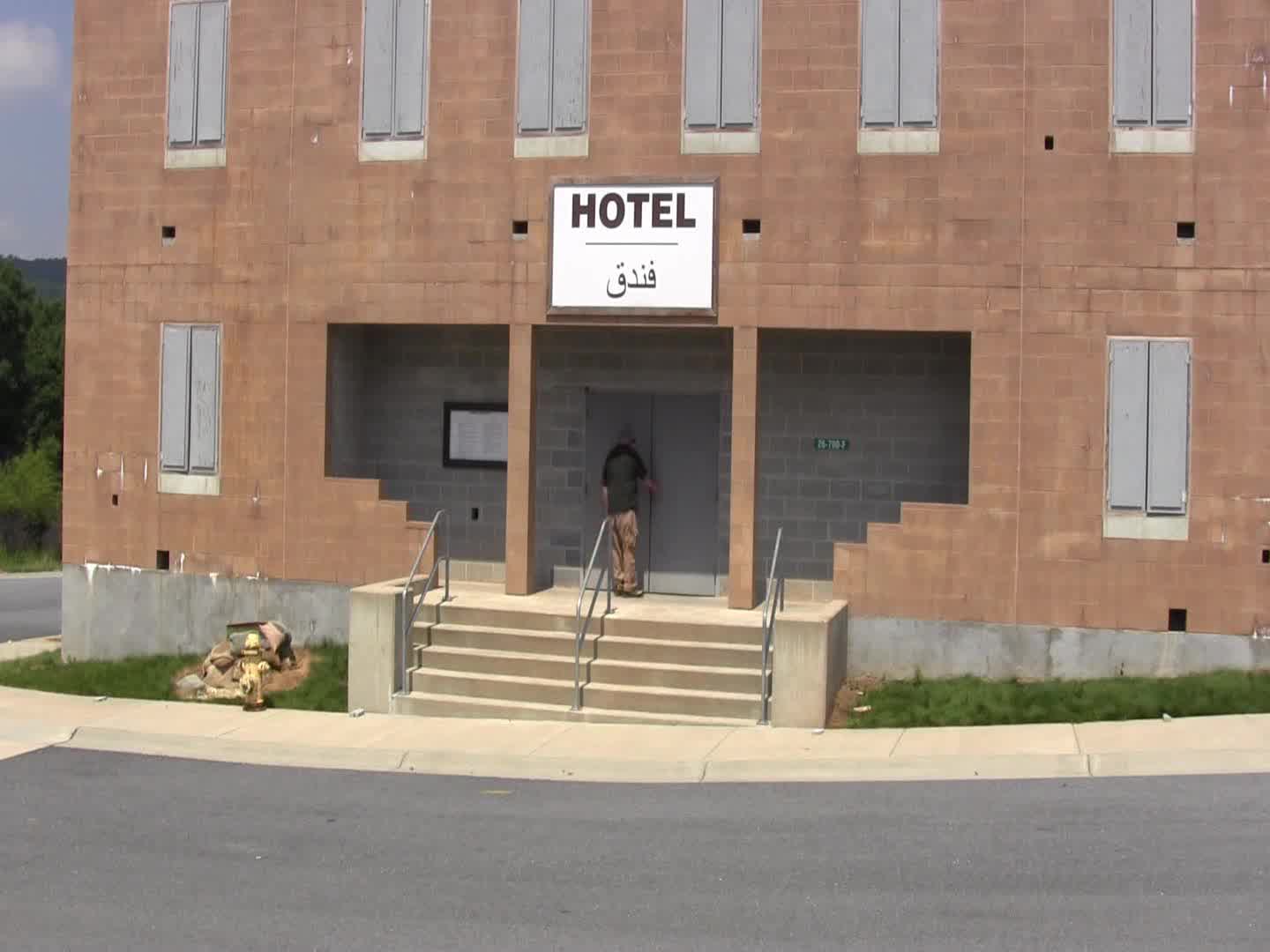}&
    \includegraphics[width=0.15\textwidth,natwidth=640,natheight=480]{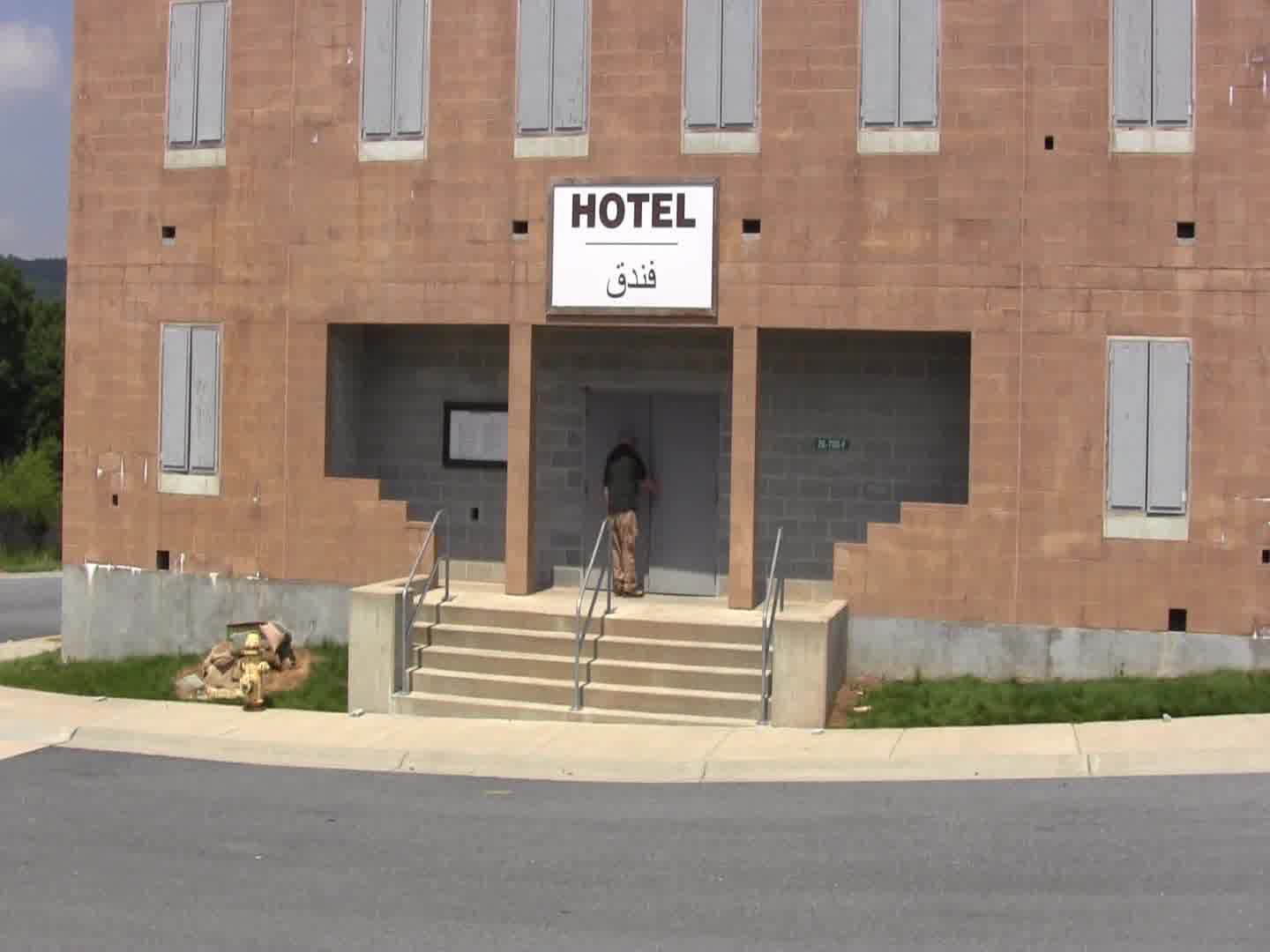}&
    \includegraphics[width=0.15\textwidth,natwidth=640,natheight=480]{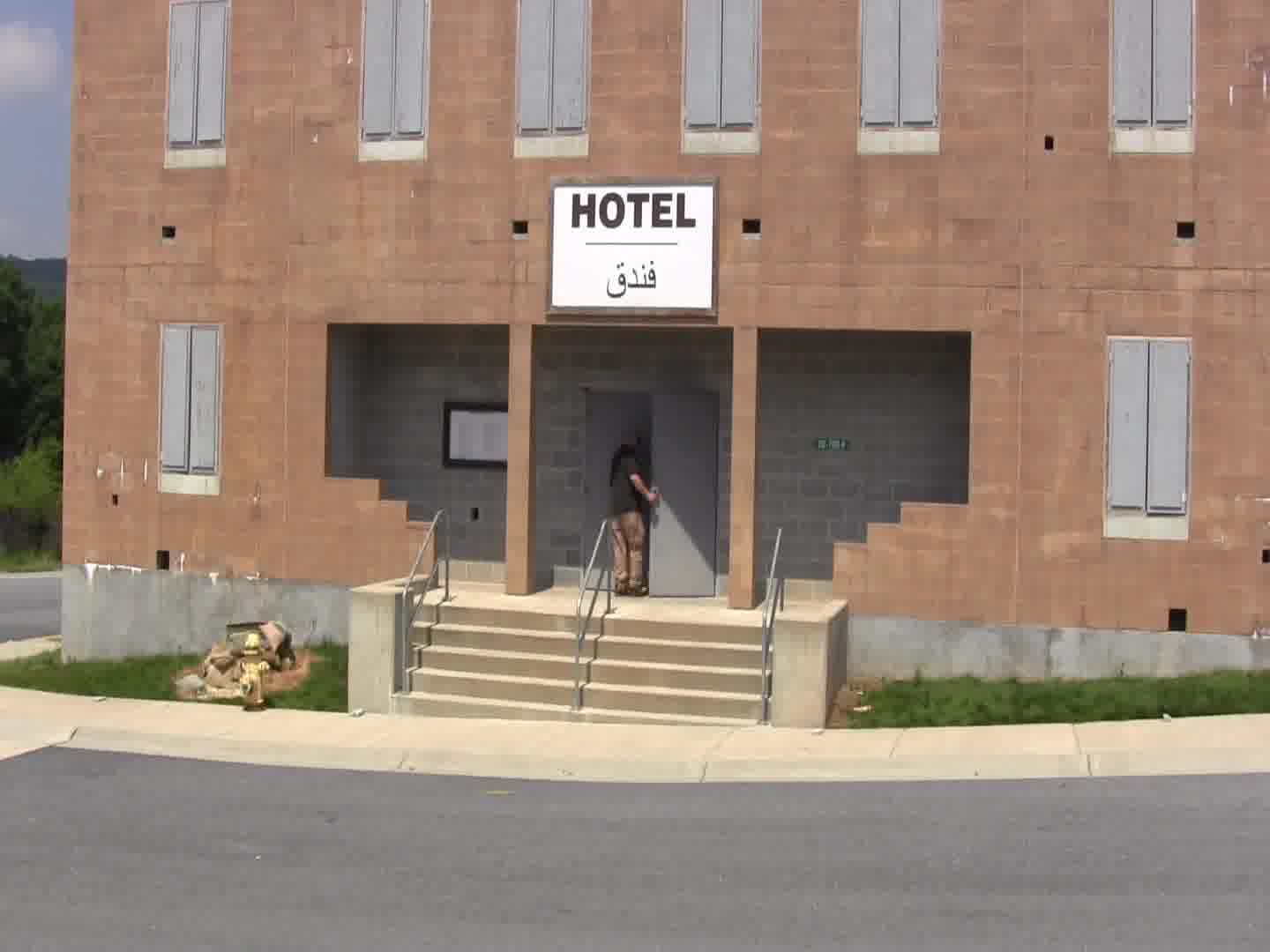}\\
    \includegraphics[width=0.15\textwidth,natwidth=640,natheight=480]{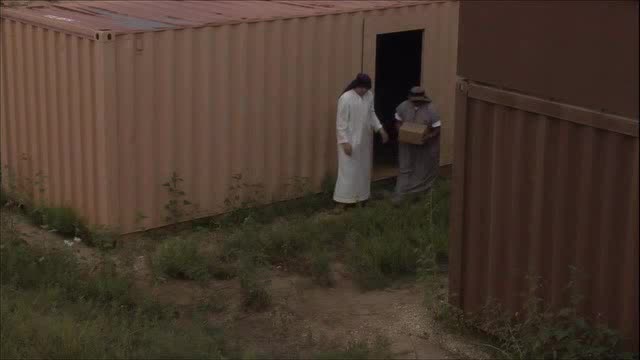}&
    \includegraphics[width=0.15\textwidth,natwidth=640,natheight=480]{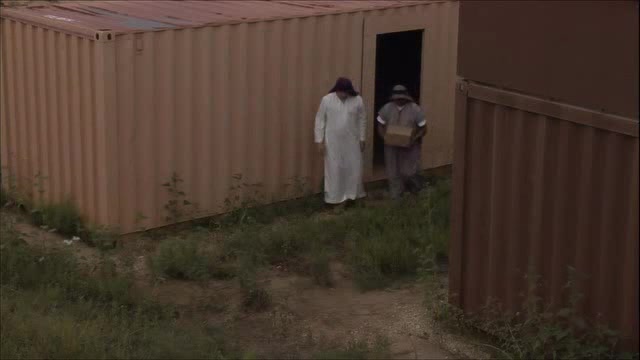}&
    \includegraphics[width=0.15\textwidth,natwidth=640,natheight=480]{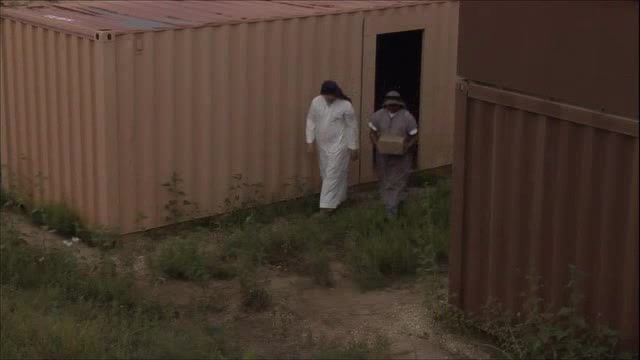}&
    \includegraphics[width=0.15\textwidth,natwidth=640,natheight=480]{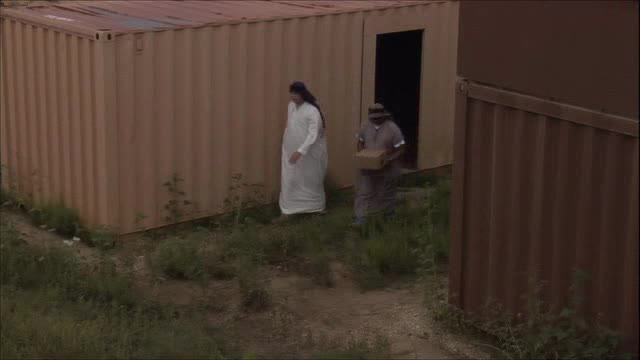}&
    \includegraphics[width=0.15\textwidth,natwidth=640,natheight=480]{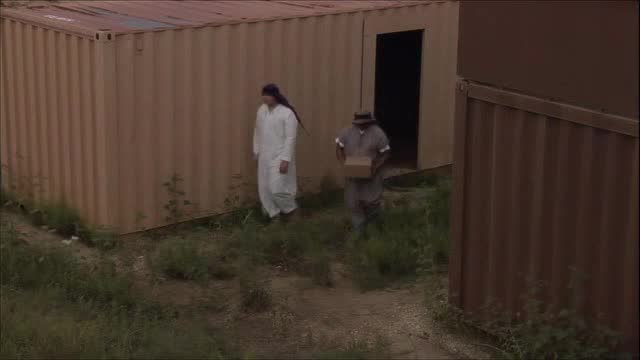}&
    \includegraphics[width=0.15\textwidth,natwidth=640,natheight=480]{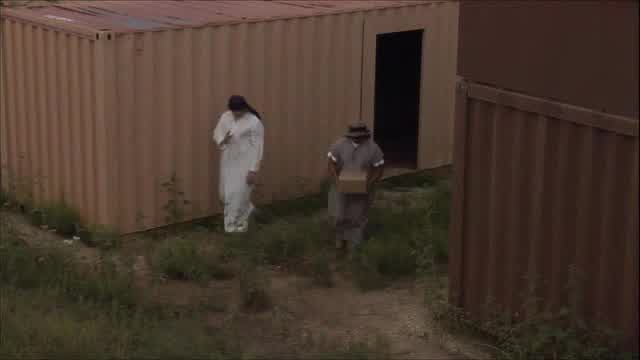}\\
    \includegraphics[width=0.15\textwidth,natwidth=640,natheight=480]{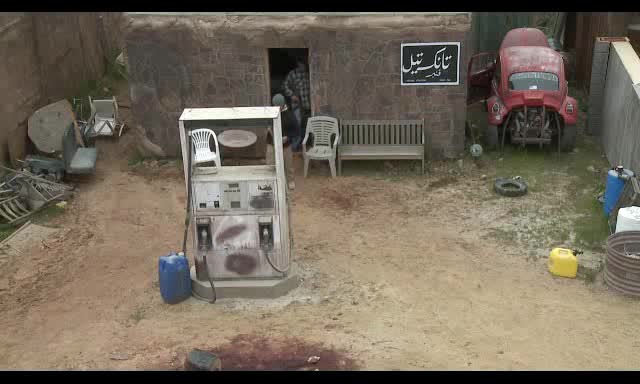}&
    \includegraphics[width=0.15\textwidth,natwidth=640,natheight=480]{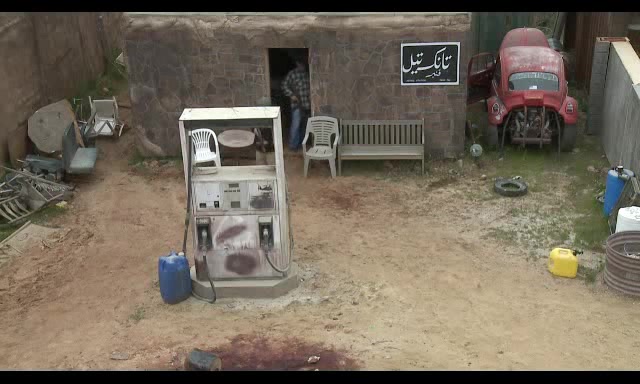}&
    \includegraphics[width=0.15\textwidth,natwidth=640,natheight=480]{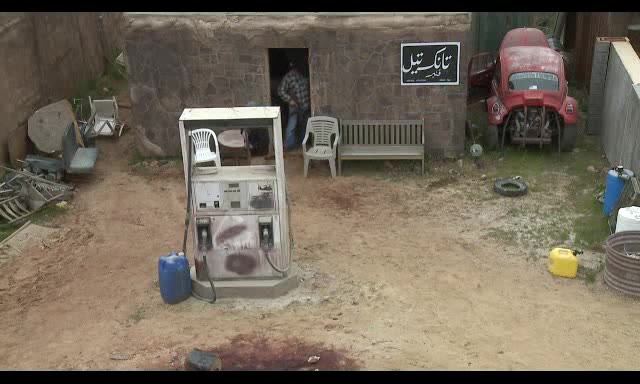}&
    \includegraphics[width=0.15\textwidth,natwidth=640,natheight=480]{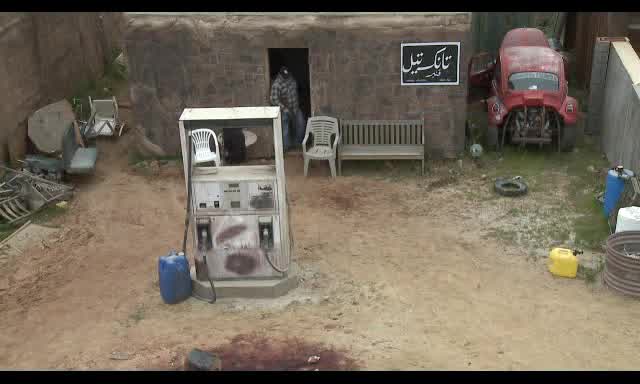}&
    \includegraphics[width=0.15\textwidth,natwidth=640,natheight=480]{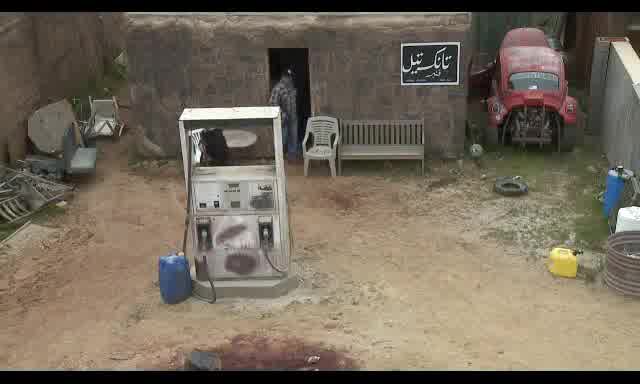}&
    \includegraphics[width=0.15\textwidth,natwidth=640,natheight=480]{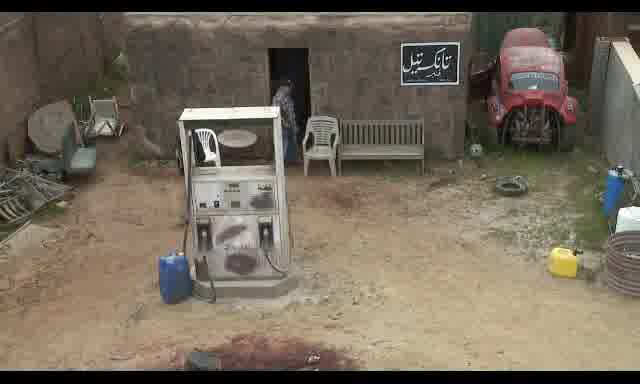}\\
    \includegraphics[width=0.15\textwidth,natwidth=640,natheight=480]{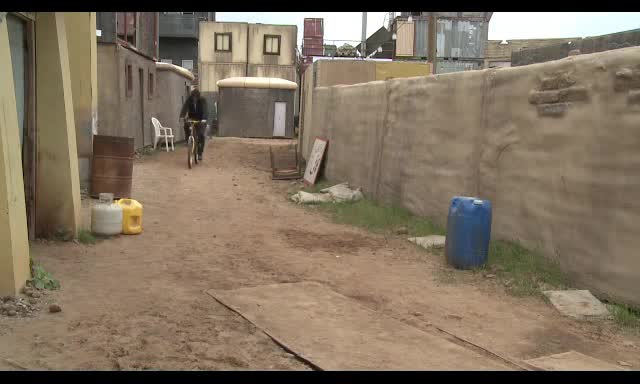}&
    \includegraphics[width=0.15\textwidth,natwidth=640,natheight=480]{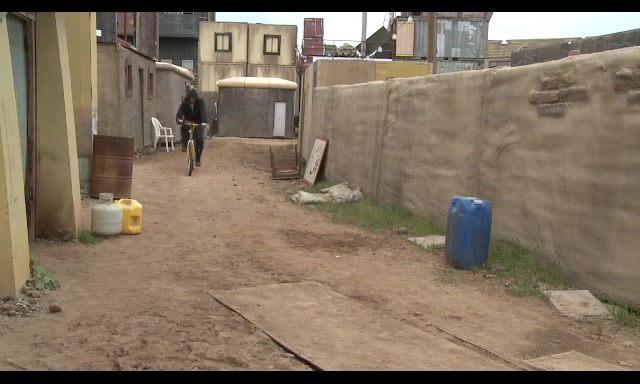}&
    \includegraphics[width=0.15\textwidth,natwidth=640,natheight=480]{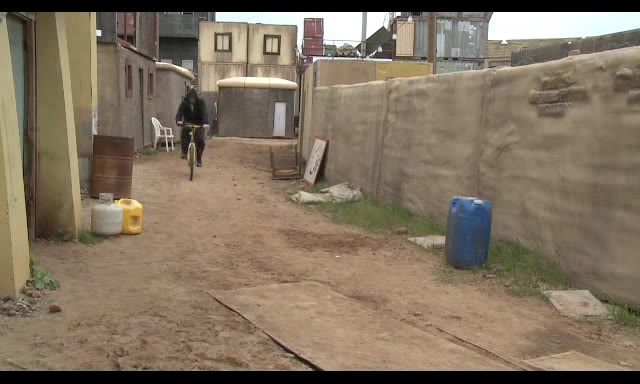}&
    \includegraphics[width=0.15\textwidth,natwidth=640,natheight=480]{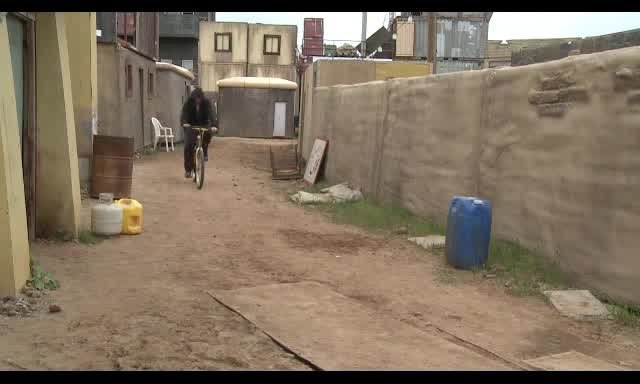}&
    \includegraphics[width=0.15\textwidth,natwidth=640,natheight=480]{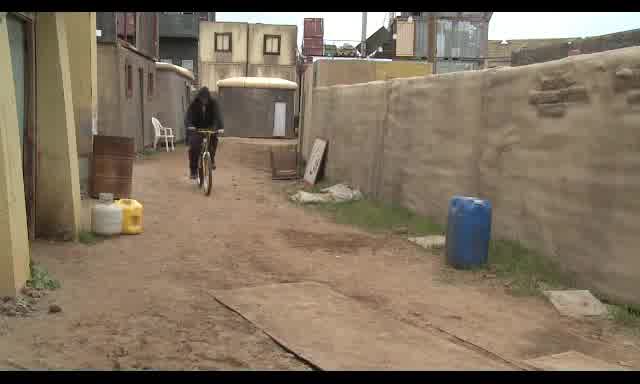}&
    \includegraphics[width=0.15\textwidth,natwidth=640,natheight=480]{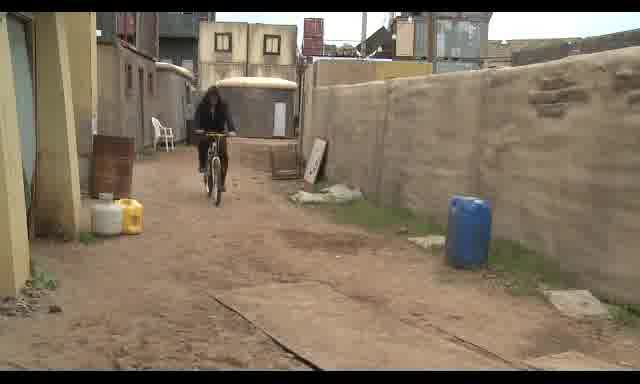}\\
    \includegraphics[width=0.15\textwidth,natwidth=640,natheight=480]{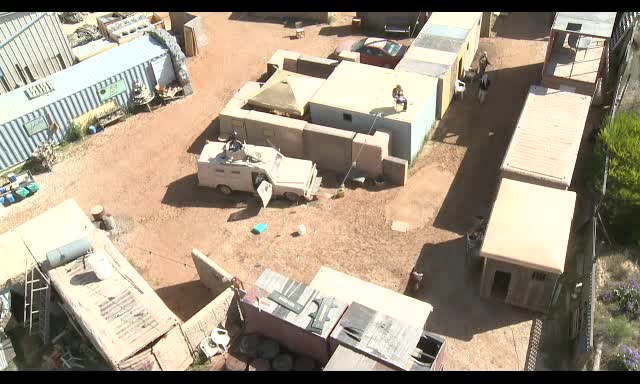}&
    \includegraphics[width=0.15\textwidth,natwidth=640,natheight=480]{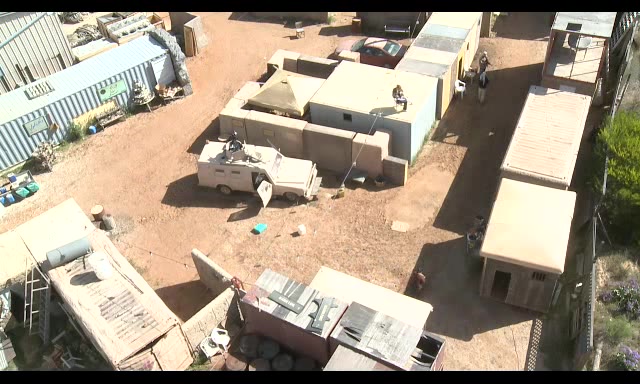}&
    \includegraphics[width=0.15\textwidth,natwidth=640,natheight=480]{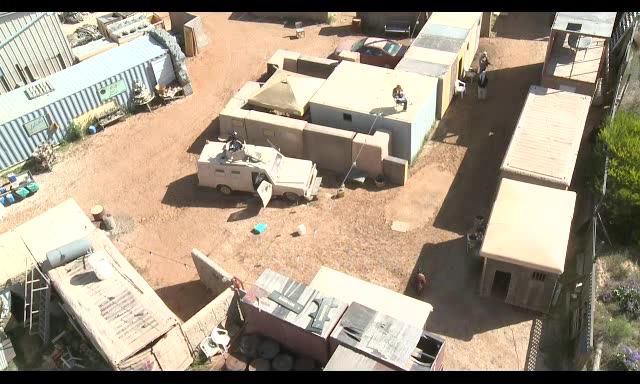}&
    \includegraphics[width=0.15\textwidth,natwidth=640,natheight=480]{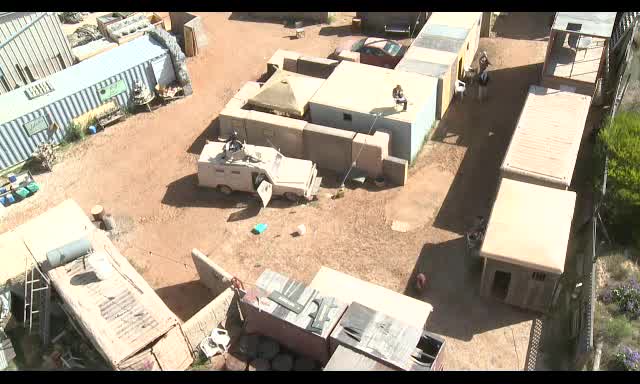}&
    \includegraphics[width=0.15\textwidth,natwidth=640,natheight=480]{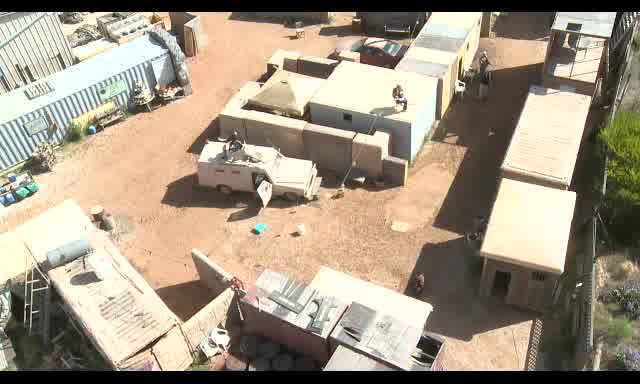}&
    \includegraphics[width=0.15\textwidth,natwidth=640,natheight=480]{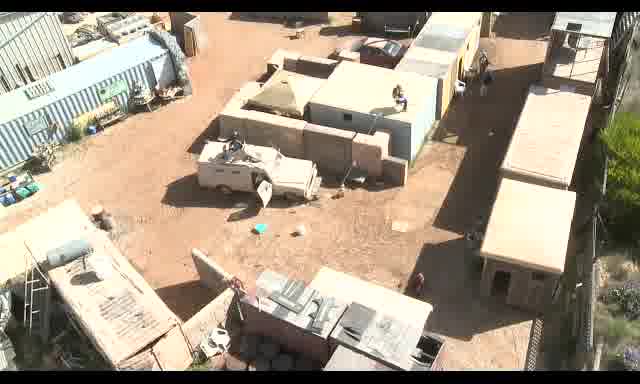}\\
    \includegraphics[width=0.15\textwidth,natwidth=640,natheight=480]{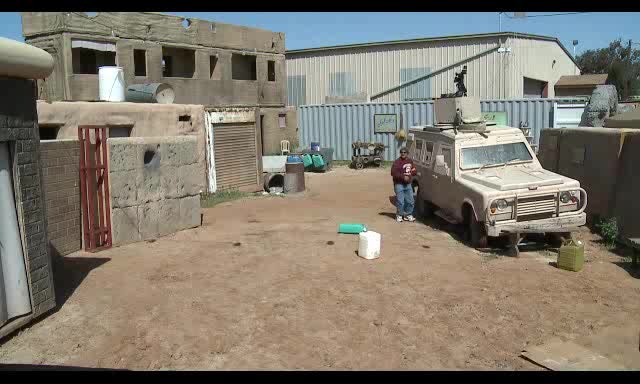}&
    \includegraphics[width=0.15\textwidth,natwidth=640,natheight=480]{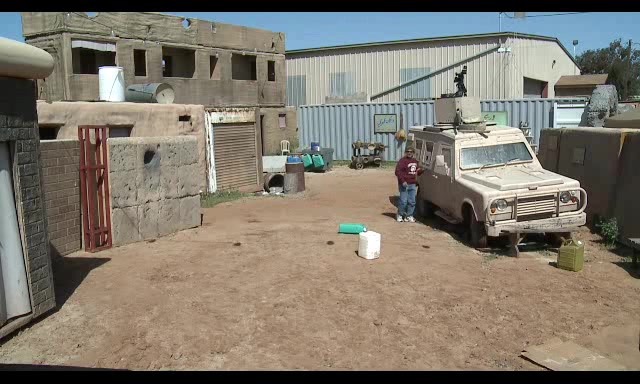}&
    \includegraphics[width=0.15\textwidth,natwidth=640,natheight=480]{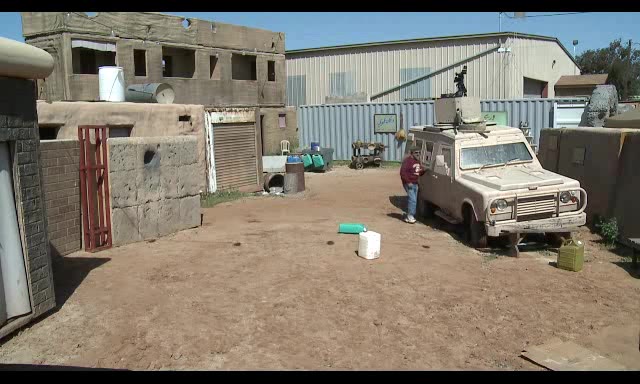}&
    \includegraphics[width=0.15\textwidth,natwidth=640,natheight=480]{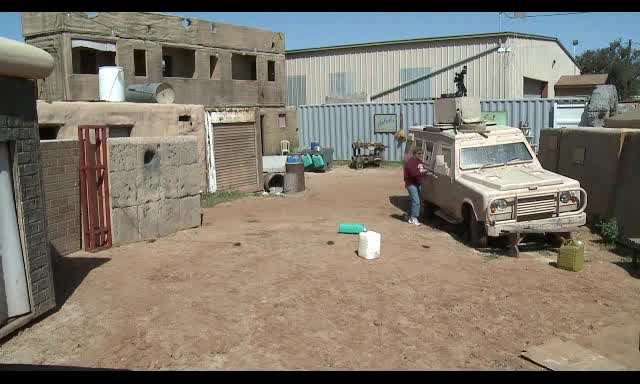}&
    \includegraphics[width=0.15\textwidth,natwidth=640,natheight=480]{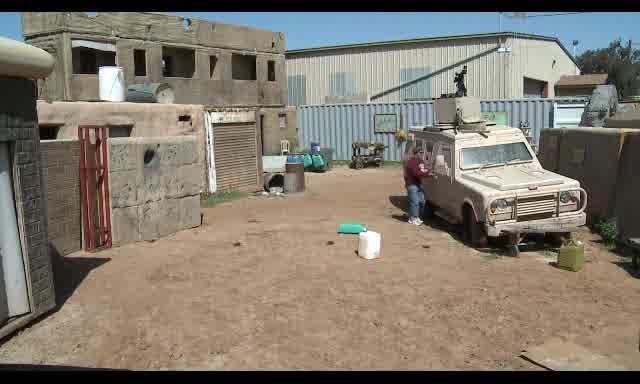}&
    \includegraphics[width=0.15\textwidth,natwidth=640,natheight=480]{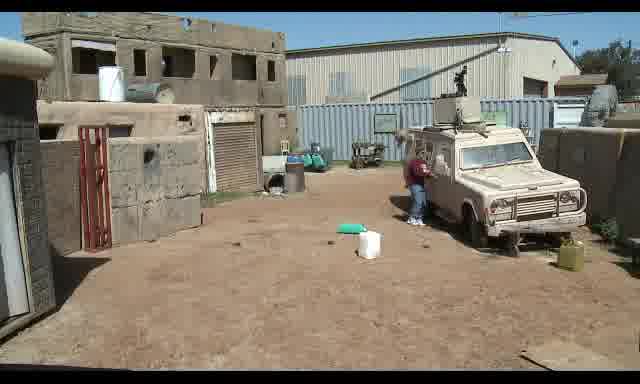}\\
    \includegraphics[width=0.15\textwidth,natwidth=640,natheight=480]{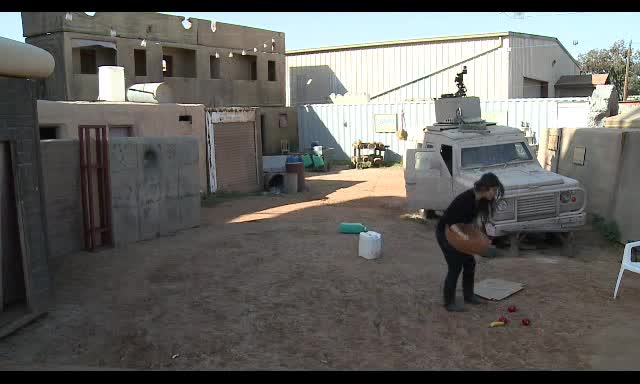}&
    \includegraphics[width=0.15\textwidth,natwidth=640,natheight=480]{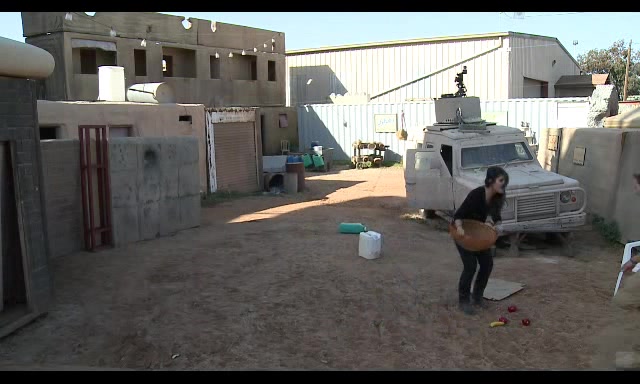}&
    \includegraphics[width=0.15\textwidth,natwidth=640,natheight=480]{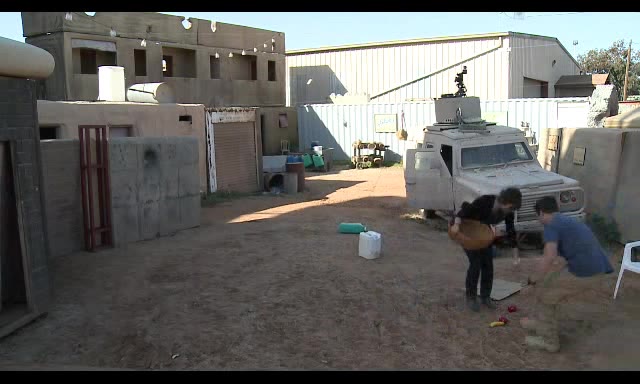}&
    \includegraphics[width=0.15\textwidth,natwidth=640,natheight=480]{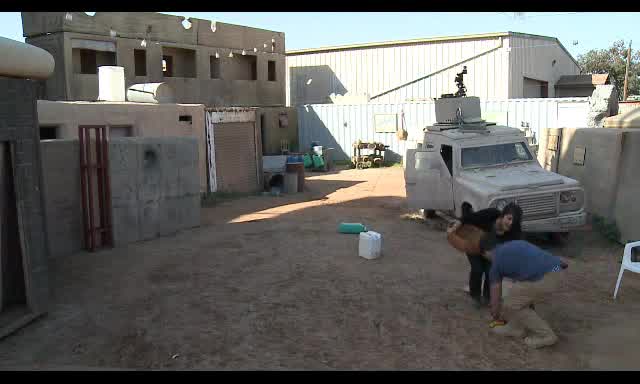}&
    \includegraphics[width=0.15\textwidth,natwidth=640,natheight=480]{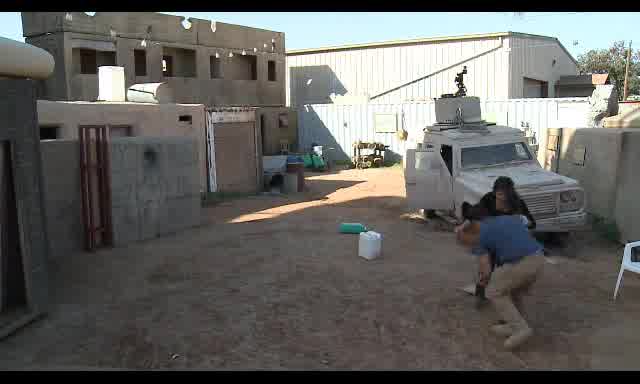}&
    \includegraphics[width=0.15\textwidth,natwidth=640,natheight=480]{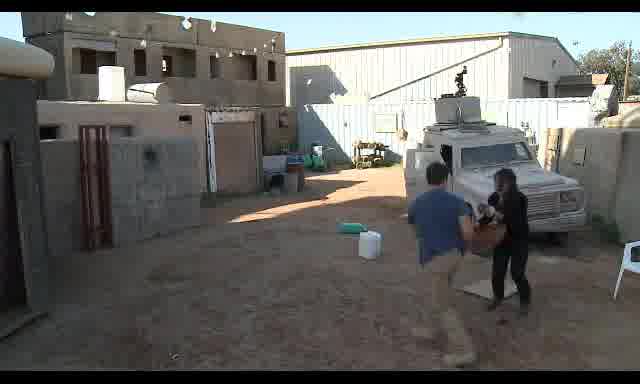}\\
    \includegraphics[width=0.15\textwidth,natwidth=640,natheight=480]{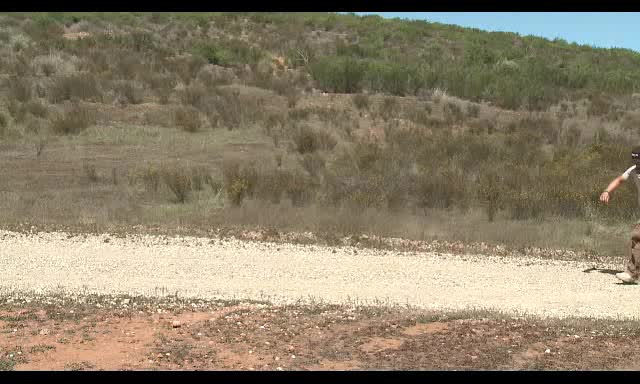}&
    \includegraphics[width=0.15\textwidth,natwidth=640,natheight=480]{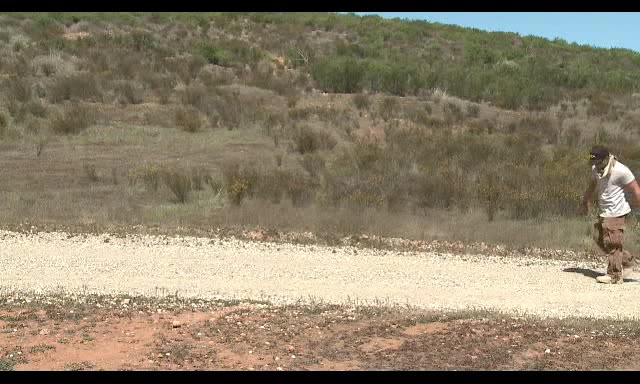}&
    \includegraphics[width=0.15\textwidth,natwidth=640,natheight=480]{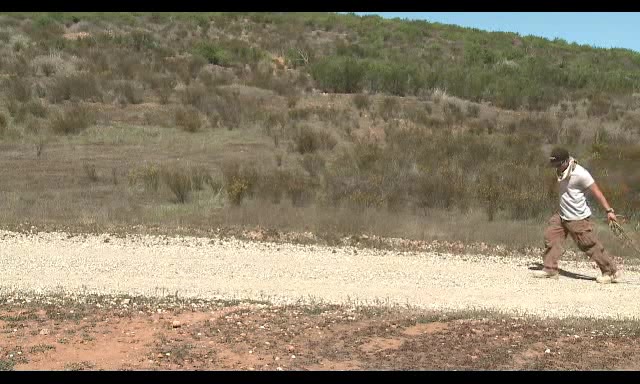}&
    \includegraphics[width=0.15\textwidth,natwidth=640,natheight=480]{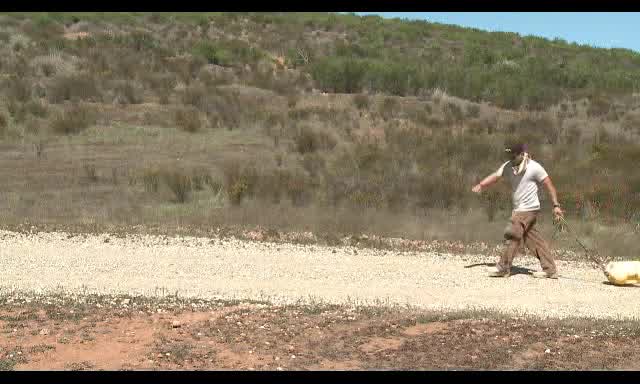}&
    \includegraphics[width=0.15\textwidth,natwidth=640,natheight=480]{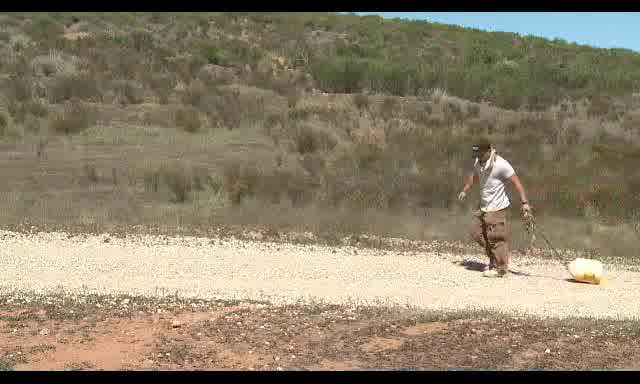}&
    \includegraphics[width=0.15\textwidth,natwidth=640,natheight=480]{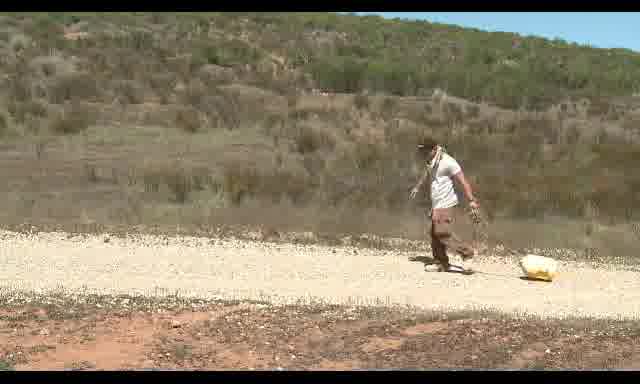}\\
    \includegraphics[width=0.15\textwidth,natwidth=640,natheight=480]{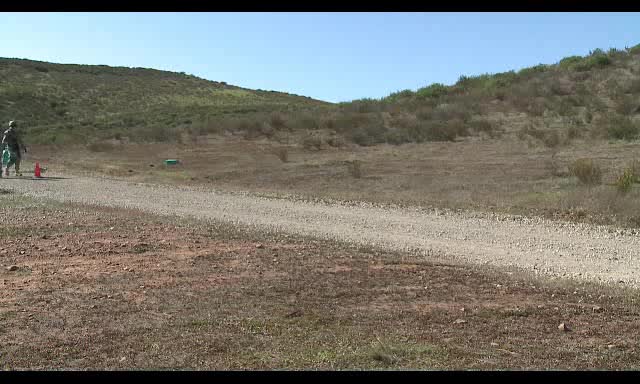}&
    \includegraphics[width=0.15\textwidth,natwidth=640,natheight=480]{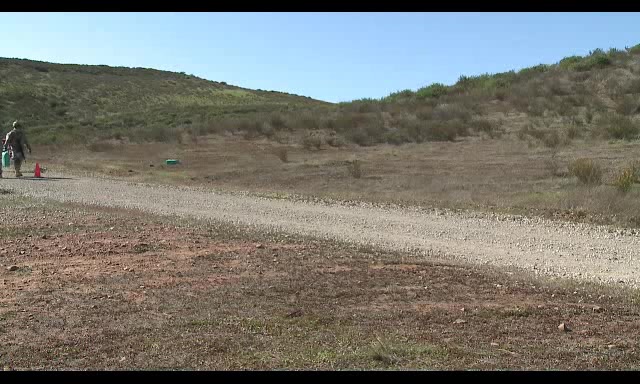}&
    \includegraphics[width=0.15\textwidth,natwidth=640,natheight=480]{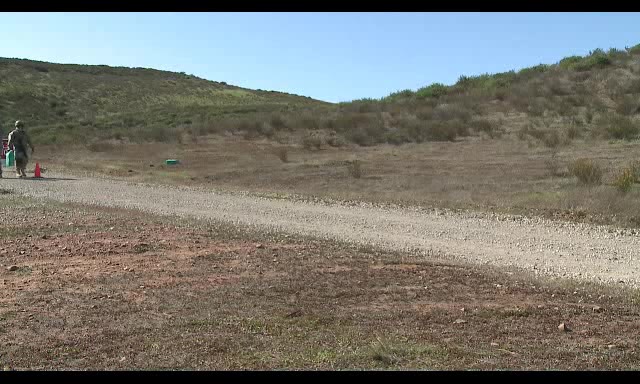}&
    \includegraphics[width=0.15\textwidth,natwidth=640,natheight=480]{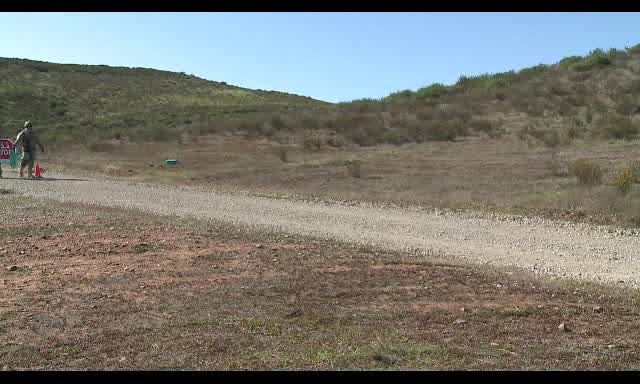}&
    \includegraphics[width=0.15\textwidth,natwidth=640,natheight=480]{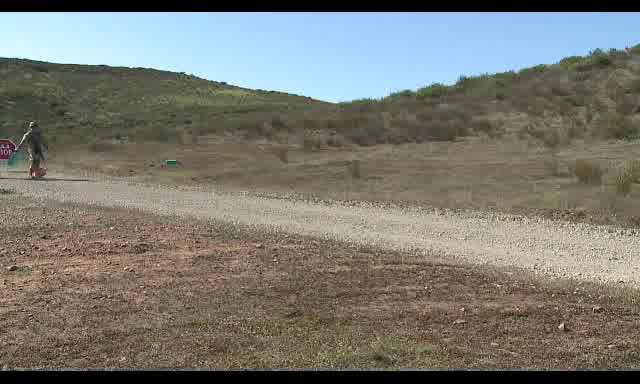}&
    \includegraphics[width=0.15\textwidth,natwidth=640,natheight=480]{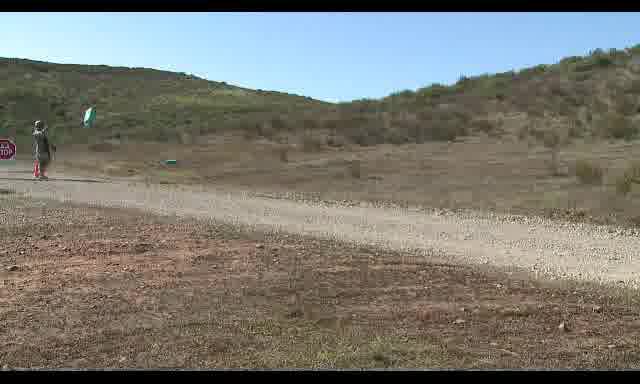}\\
    \includegraphics[width=0.15\textwidth,natwidth=640,natheight=480]{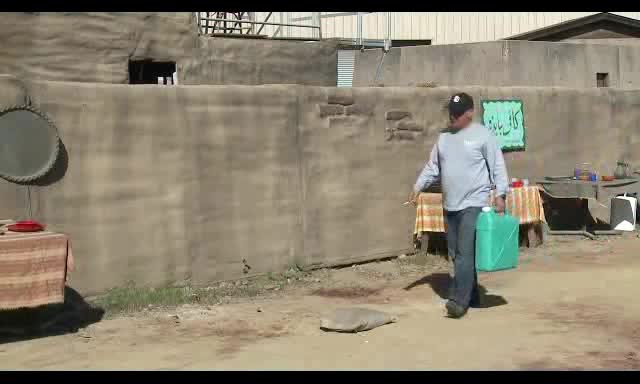}&
    \includegraphics[width=0.15\textwidth,natwidth=640,natheight=480]{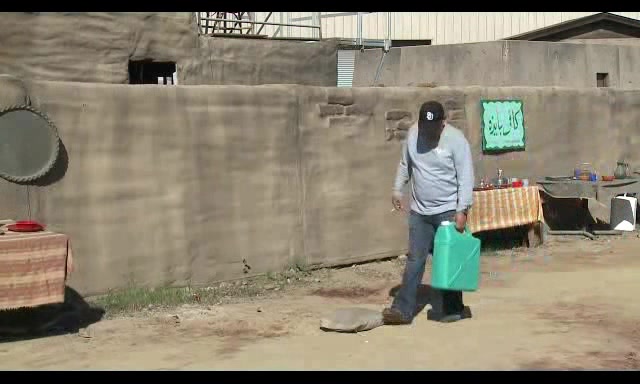}&
    \includegraphics[width=0.15\textwidth,natwidth=640,natheight=480]{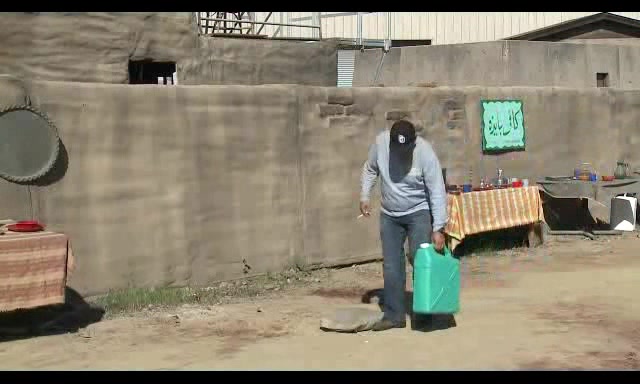}&
    \includegraphics[width=0.15\textwidth,natwidth=640,natheight=480]{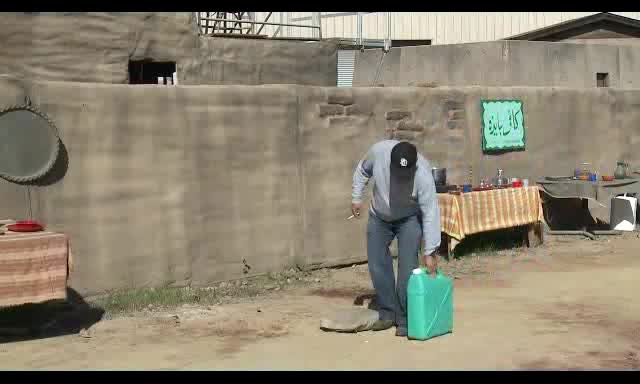}&
    \includegraphics[width=0.15\textwidth,natwidth=640,natheight=480]{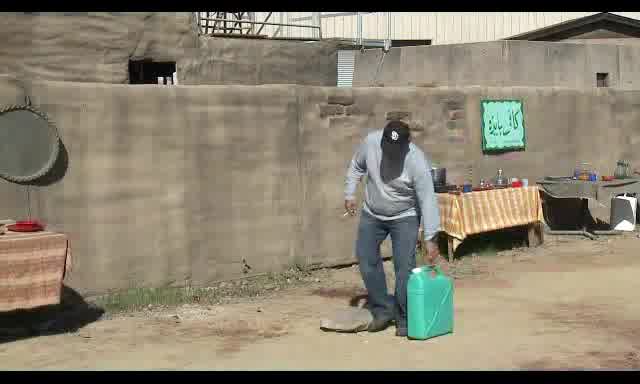}&
    \includegraphics[width=0.15\textwidth,natwidth=640,natheight=480]{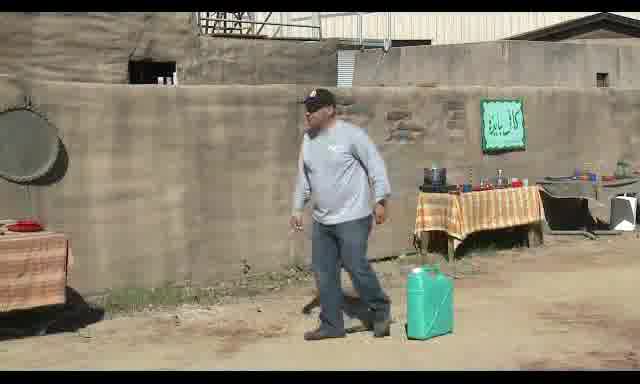}
  \end{tabular}
  \caption{Several frame sequences from the LCA dataset illustrating several of
    the backgrounds in which they were filmed.}
  \label{fig:frames}
\end{figure*}

A portion of the video was annotated by Mitre for the Mind's Eye Year~2
evaluation.
That annotation is not included in LCA.\@
After the completion of the Mind's Eye Year~2 evaluation, we undertook a
systematic annotation effort for a portion of the above video.
That annotation forms the basis of LCA.\@
LCA contains all and only the portion of the above video that was annotated as
part of this process.
%
%
This video comprises 190 files as delineated in Table~\ref{tab:files}.
Eight files are MOV format, 46~are MP4 format, and 136~are AVI format.
The MOV files all use the MP4V codec and are 640$\times$384 at 60~fps.
The MP4 files all use the H264 codec and are 640$\times$360 at 30~fps.
The AVI files all use the XVID codec and are either 640$\times$384 at 60~fps
or 1440$\times$1080 at 30~fps.
This constitutes 2302144 frames and a total of 12~hours, 51~minutes, and
16~seconds of video.
%
%
For comparison, UCF50 has 1330936 frames and 13.81~hours, HMDB51 has 632635
frames and 5.85~hours, UCF101 has 27~hours, Hollywood-2 has 20.1~hours, and
VIRAT has 8.5~hours.
Several frame sequences from this dataset illustrating several of the
backgrounds are shown in Fig.~\ref{fig:frames}.

\setcounter{table}{1}

\begin{table*}
  \caption{
    The original names of the files provided by DARPA.\@
    Filenames containing GPTC were filmed at GPJTC.\@
    Filenames containing STOPS were filmed at STOPS.\@
    Filenames consisting solely of a number were filmed at FITG.\@
    Numbers of the form YYYYMMDD indicate filming date.
    CR indicates country road.
    SH indicates safe house.
    Indices on CR, SH, and VT indicate variant backgrounds of the given class.
    CP1, CP2, C1, and C3 indicate camera.
    Text indicates the staging directions to guide filming.
    The remaining numbers serve to uniquely identify the video.
    These videos were renamed to \texttt{video-XXX} for consistency in the
    release.
    The release also contains a file, \texttt{video-mapping.txt}, which includes
    the mapping between the original filenames and those in the LCA release.}
  \label{tab:files}
  \centering
  \begin{tt}
    \resizebox{\textwidth}{!}{\begin{tabular}{@{}cccc@{}}
      \begin{tabular}[t]{@{}l@{}}
        GPTC\_20110822\_SH\_02\_CP1\_NOACTIVITY\\
        GPTC\_20110822\_SH\_02\_CP2\_NOACTIVITY\\
        GPTC\_20110822\_SH\_07\_CP1\_EX\&RET\\
        GPTC\_20110822\_SH\_07\_CP2\_EX\&RET\\
        GPTC\_20110822\_SH\_13\_CP1\_WARYHO\\
        GPTC\_20110822\_SH\_13\_CP2\_WARYHO\\
        GPTC\_20110822\_SH\_11\_CP1\_PARAHO\\
        GPTC\_20110822\_SH\_11\_CP2\_PARAHO\\
        GPTC\_20110822\_SH\_12\_CP1\_HOSTILEHO\\
        GPTC\_20110822\_SH\_12\_CP2\_HOSTILEHO\\
        GPTC\_20110822\_SH\_06\_CP1\_SUPCACHRET\\
        GPTC\_20110822\_SH\_06\_CP2\_SUPCACHRET\\
        GPTC\_20110822\_SH\_09\_CP1\_GIVE\&CONT-BLDG\\
        GPTC\_20110822\_SH\_09\_CP2\_GIVE\&CONT-BLDG\\
        GPTC\_20110822\_SH\_05\_CP1\_SUPCACHDUMP\\
        GPTC\_20110822\_SH\_05\_CP2\_SUPCACHDUMP\\
        GPTC\_20110822\_SH\_08\_CP1\_HO\&EX-mixBKG\\
        GPTC\_20110822\_SH\_08\_CP2\_HO\&EX-mixBKG\\
        GPTC\_20110822\_SH\_03\_CP1\_HO\&RET\\
        GPTC\_20110822\_SH\_03\_CP2\_HO\&RET\\
        GPTC\_20110822\_SH\_04\_CP1\_HO\&RET2\\
        GPTC\_20110822\_SH\_04\_CP2\_HO\&RET2\\
        GPTC\_20110822\_CR\_13\_CP1\_NOACTIVITY\\
        GPTC\_20110822\_CR\_13\_CP2\_NOACTIVITY\\
        GPTC\_20110822\_CR\_07\_CP1\_SALESMAN\\
        GPTC\_20110822\_CR\_07\_CP2\_SALESMAN\\
        GPTC\_20110822\_CR\_12\_CP1\_BAGDOWNHO\&RET\\
        GPTC\_20110822\_CR\_12\_CP2\_BAGDOWNHO\&RET\\
        GPTC\_20110822\_CR\_02\_CP1\_RoutineActivity\\
        GPTC\_20110822\_CR\_02\_CP2\_RoutineActivity\\
        GPTC\_20110822\_CR\_05\_CP1\_CONTHUR-NOHO\\
        GPTC\_20110822\_CR\_05\_CP2\_CONTHUR-NOHO\\
        GPTC\_20110822\_CR\_04\_CP1\_EX\\
        GPTC\_20110822\_CR\_04\_CP2\_EX\\
        GPTC\_20110822\_CR\_08\_CP1\_SALESMAN\_INSIST\\
        GPTC\_20110822\_CR\_08\_CP2\_SALESMAN\_INSIST\\
        GPTC\_20110822\_CR\_11\_CP1\_DOUBLEAVOIDBADGUY\\
        GPTC\_20110822\_CR\_11\_CP2\_DOUBLEAVOIDBADGUY\\
        GPTC\_20110822\_CR\_09\_CP1\_UPTONOGOOD\\
        GPTC\_20110822\_CR\_09\_CP2\_UPTONOGOOD\\
        GPTC\_20110822\_CR\_06\_CP1\_DISAGREE\_NOGIVE\_RETURN\\
        GPTC\_20110822\_CR\_06\_CP2\_DISAGREE\_NOGIVE\_RETURN\\
        GPTC\_20110822\_CR\_10\_CP1\_AVOIDBADGUY\\
        GPTC\_20110822\_CR\_10\_CP2\_AVOIDBADGUY\\
        GPTC\_20110822\_CR\_03\_CP1\_HO\\
        GPTC\_20110822\_CR\_03\_CP2\_HO
      \end{tabular}&
      \begin{tabular}[t]{@{}l@{}}
        STOPS\_20120307\_SH3\_06\_C1-edited-01\\
        STOPS\_20120307\_SH3\_06\_C3-edited-01\\
        STOPS\_20120307\_SH3\_03\_C1-edited-01\\
        STOPS\_20120307\_SH3\_03\_C3-edited-01\\
        STOPS\_20120307\_SH3\_02\_C1-edited-01\\
        STOPS\_20120307\_SH3\_02\_C3-edited-01\\
        STOPS\_20120307\_SH3\_01\_C1-edited-01\\
        STOPS\_20120307\_SH3\_01\_C3-edited-01\\
        STOPS\_20120308\_CR1\_07a\_C1\\
        STOPS\_20120308\_CR1\_07a\_C3\\
        STOPS\_20120309\_VT1\_23\_C1\\
        STOPS\_20120309\_VT1\_23\_C3\\
        STOPS\_20120308\_CR1\_NA\_C1\\
        STOPS\_20120308\_CR1\_NA\_C3\\
        STOPS\_20120306\_SH1\_06\_C1\\
        STOPS\_20120306\_SH1\_06\_C3\\
        STOPS\_20120306\_SH1\_NA\_C1\\
        STOPS\_20120306\_SH1\_NA\_C3\\
        STOPS\_20120308\_CR1\_08a\_C1\\
        STOPS\_20120308\_CR1\_08a\_C3\\
        STOPS\_20120308\_CR1\_12a\_C1\\
        STOPS\_20120308\_CR1\_12a\_C3\\
        STOPS\_20120308\_CR1\_01a\_C1\\
        STOPS\_20120308\_CR1\_01a\_C3\\
        STOPS\_20120309\_VT1\_01\_C1\\
        STOPS\_20120309\_VT1\_01\_C3\\
        STOPS\_20120309\_VT1\_26\_C1\\
        STOPS\_20120309\_VT1\_26\_C3\\
        STOPS\_20120308\_CR1\_02a\_C1\\
        STOPS\_20120308\_CR1\_02a\_C3\\
        STOPS\_20120308\_CR1\_05a\_C1\\
        STOPS\_20120308\_CR1\_05a\_C3\\
        STOPS\_20120309\_VT1\_04\_C1\\
        STOPS\_20120309\_VT1\_04\_C3\\
        STOPS\_20120306\_SH1\_07\_C1\\
        STOPS\_20120306\_SH1\_07\_C3\\
        STOPS\_20120309\_VT1\_21\_C1\\
        STOPS\_20120309\_VT1\_21\_C3\\
        STOPS\_20120307\_SH3\_07\_C1\\
        STOPS\_20120307\_SH3\_07\_C3\\
        STOPS\_20120308\_CR1\_09a\_C1\\
        STOPS\_20120308\_CR1\_09a\_C3\\
        STOPS\_20120307\_SH3\_08\_C1\\
        STOPS\_20120307\_SH3\_08\_C3\\
        STOPS\_20120308\_CR1\_11a\_C1\\
        STOPS\_20120308\_CR1\_11a\_C3\\
        STOPS\_20120306\_SH1\_05\_C1\\
        STOPS\_20120306\_SH1\_05\_C3\\
        STOPS\_20120309\_VT1\_12\_C1\\
        STOPS\_20120309\_VT1\_12\_C3\\
        STOPS\_20120309\_VT1\_10\_C1\\
        STOPS\_20120309\_VT1\_10\_C3
      \end{tabular}&
      \begin{tabular}[t]{@{}l@{}}
        STOPS\_20120308\_CR1\_06a\_C1\\
        STOPS\_20120308\_CR1\_06a\_C3\\
        STOPS\_20120306\_SH1\_04\_C1\\
        STOPS\_20120306\_SH1\_04\_C3\\
        STOPS\_20120306\_SH1\_01\_C1\\
        STOPS\_20120306\_SH1\_01\_C3\\
        STOPS\_20120308\_CR1\_13a\_C1\\
        STOPS\_20120308\_CR1\_13a\_C3\\
        STOPS\_20120309\_VT1\_07\_C1\\
        STOPS\_20120309\_VT1\_07\_C3\\
        STOPS\_20120309\_VT1\_25\_C1\\
        STOPS\_20120309\_VT1\_25\_C3\\
        STOPS\_20120309\_VT1\_05\_C1\\
        STOPS\_20120309\_VT1\_05\_C3\\
        STOPS\_20120308\_CR1\_01c\_C1\\
        STOPS\_20120308\_CR1\_01c\_C3\\
        STOPS\_20120309\_VT1\_NA\_C1\\
        STOPS\_20120309\_VT1\_NA\_C3\\
        STOPS\_20120308\_CR1\_03a\_C1\\
        STOPS\_20120308\_CR1\_03a\_C3\\
        STOPS\_20120306\_SH1\_02\_C1\\
        STOPS\_20120306\_SH1\_02\_C3\\
        STOPS\_20120307\_SH3\_11\_C1\\
        STOPS\_20120307\_SH3\_11\_C3\\
        STOPS\_20120308\_CR1\_04a\_C1\\
        STOPS\_20120308\_CR1\_04a\_C3\\
        STOPS\_20120307\_SH3\_10\_C1\\
        STOPS\_20120307\_SH3\_10\_C3\\
        STOPS\_20120309\_VT1\_22\_C1\\
        STOPS\_20120309\_VT1\_22\_C3\\
        STOPS\_20120308\_CR1\_01b\_C1\\
        STOPS\_20120308\_CR1\_01b\_C3\\
        STOPS\_20120309\_VT1\_28\_C1\\
        STOPS\_20120309\_VT1\_28\_C3\\
        STOPS\_20120309\_VT1\_20\_C1\\
        STOPS\_20120309\_VT1\_20\_C3\\
        STOPS\_20120309\_VT1\_03\_C1\\
        STOPS\_20120309\_VT1\_03\_C3\\
        STOPS\_20120309\_VT1\_27\_C1\\
        STOPS\_20120309\_VT1\_27\_C3\\
        STOPS\_20120309\_VT1\_13\_C1\\
        STOPS\_20120309\_VT1\_13\_C3\\
        STOPS\_20120308\_CR1\_06b\_C1\\
        STOPS\_20120308\_CR1\_06b\_C3\\
        STOPS\_20120309\_VT1\_02\_C1\\
        STOPS\_20120309\_VT1\_02\_C3\\
        STOPS\_20120308\_CR1\_10a\_C1\\
        STOPS\_20120308\_CR1\_10a\_C3\\
        STOPS\_20120309\_VT1\_24\_C1\\
        STOPS\_20120309\_VT1\_24\_C3\\
        STOPS\_20120308\_CR1\_03b\_C1\\
        STOPS\_20120308\_CR1\_03b\_C3
      \end{tabular}&
      \begin{tabular}[t]{@{}l@{}}
        100\\
        105\\
        114\\
        118\\
        11\\
        121\\
        128\\
        133\\
        154\\
        158\\
        162\\
        175\\
        188\\
        195\\
        196\\
        200\\
        211\\
        226\\
        22\\
        230\\
        238\\
        243\\
        252\\
        265\\
        269\\
        277\\
        30\\
        39\\
        41\\
        46\\
        52\\
        57\\
        59\\
        62\\
        68\\
        69\\
        81\\
        86\\
        88\\
        97
      \end{tabular}
    \end{tabular}}
  \end{tt}
\end{table*}

The Mind's Eye program specified a set of 48~verbs of interest.
Of these, the LCA dataset uses only 24~verbs as annotation labels, as delineated
in Table~\ref{tab:verbs}.
Of these, 17~verbs were used as part of the stage directions given to the
actors to guide the actions that they performed.
The remainder were not used as part of the stage directions but occurred
incidentally.
Nothing, however, precluded the actors from performing actions that could be
described by other verbs.
Thus the video depicts many other actions than those annotated, including but
not limited to riding bicycles, pushing carts, singing, pointing guns, arguing,
and kicking balls.
The only restriction, in principle, to these 24~verbs is that these were the
only actions that were annotated.
Identifying the presence of specific verbs in the context of many such
confounding actions should present additional challenges.

\setcounter{table}{0}

\begin{table}
\caption{Verbs used as labels in the LCA dataset.
    The starred verbs were used as part of the stage directions to the actors.
    The remaining verbs were not used as part of the stage directions but may
    have occurred incidentally.}
  \label{tab:verbs}
  \centering
  \begin{em}
    \begin{tabular}{@{}llll@{}}
      approach$^{*}$&
      drop$^{*}$&
      give$^{*}$&
      replace$^{*}$\\
      arrive&
      enter$^{*}$&
      hold&
      run\\
      bury$^{*}$&
      exchange$^{*}$&
      leave&
      stop\\
      carry$^{*}$&
      exit$^{*}$&
      pass$^{*}$&
      take$^{*}$\\
      chase$^{*}$&
      flee$^{*}$&
      pick up$^{*}$&
      turn\\
      dig$^{*}$&
      follow$^{*}$&
      put down$^{*}$&
      walk
    \end{tabular}
  \end{em}
\end{table}

\setcounter{table}{2}

\section{Annotation}
\label{sec:annotation}

We annotated all occurrences of the 24~verbs from Table~\ref{tab:verbs} in the
videos in Table~\ref{tab:files}.
Each such occurrence consisted of a temporal interval labeled with a verb.
The judgment of whether an action described by a particular verb occurred is
subjective; different annotators will arrive at different judgments as to
occurrence as well as the temporal extent thereof.
To help guide annotators, we gave them the specification of the intended
meaning of each of the 24~verbs as provided by DARPA.\@
Annotators performed the annotation at workstations with dual monitors.
One monitor displayed the annotation tool while the other monitor displayed the
documentation of intended verb meaning.
The documentation of intended verb meaning is included in the LCA distribution.

We also asked annotators to annotate intervals where certain object classes
were present in the field of view.
These include \emph{bandannas}, \emph{bicycles}, \emph{people}, \emph{vehicles},
and \emph{weapons}.
(The \emph{bandannas} were worn by \emph{people} around their head or arms.)
For these, a count of the number of instances of each class that were visible
in the field of view was maintained.
It was incremented each time a new instance became visible and decremented each
time an instance became invisible.
We instructed annotators that there was no need to be precise when an instance
was partially visible.
We further instructed annotators that \emph{vehicles} denoted motor vehicles,
not push carts, and \emph{weapons} denoted guns, not other things like clubs or
rocks that could be used as weapons.

We provided annotators with a tool that allowed them to view the videos at
ordinary frame rate, stop and start the videos at will, navigate to arbitrary
points in the videos, view individual frames of the videos, add, delete,
and move starting and ending points of intervals, and label intervals with
verbs.
The tool also contained buttons to increment and decrement the counts for each
of the object classes and appraised the annotator with the running counts for
the object classes in each frame as the video was played or navigated.

Because of the large quantity of video to be annotated, and the fact that
nothing happens during large portions of the video, we preprocessed the video
to reduce the amount requiring manual annotation.
We first downsampled the video to 5~fps just for the purpose of annotation;
the annotation was converted back at the end to the original frame rate.
Then segments of this downsampled video where no motion occurred were removed.
To do this, we computed dense optical flow on each pixel of each frame of the
downsampled video.
We then computed the average of the magnitude of the flow vectors in each
frame and determined which frames were above a threshold.
Stretches of contiguous frames that were above threshold that were separated by
short stretches of contiguous frame that were below threshold were merged into
single temporal segments.
Then such single temporal segments that were shorter than a specified temporal
length were discarded.\footnote{The threshold for average optical flow
  magnitude was~150.
  The threshold for ignoring short below-threshold spans when merging contiguous
  above-threshold frames into temporal segments was 50~frames.
  The threshold for ignoring short temporal segments was 15~frames.}
Annotators were only given the remaining temporal segments to annotate.
We performed a postprocessing step whereby the authors manually viewed all
discarded frames to make sure that no actions started, ended, or spanned the
omitted temporal segments.
As part of this post processing step, the authors manually checked that none of
the specified object classes entered or left the field of view during the
omitted temporal segments.

We had five annotators each independently annotate the entire LCA dataset.
Annotators were given initial instructions.
During the annotation, annotators were encouraged to discuss their annotation
judgments with the authors.
The authors would then arbitrate the judgment, often specifying principles to
guide the annotation.
These principles were then circulated among the other annotators.
The annotator instructions and principles developed through arbitration are
included in the LCA distribution.

We performed a consistency check during the annotation process.
Whenever an annotator completed annotation of a temporal segment, if that
annotator did not annotate any intervals during that segment but other
annotators did, we asked that annotator to review their annotation.

The LCA dataset contains five verb-annotation files for each of the video
files in Table~\ref{tab:files}.
These have the same name as their corresponding video, but with the extension
\texttt{txt}, and are located in directories named with each of the annotator
codes \texttt{bmedikon}, \texttt{cbushman}, \texttt{kim861}, \texttt{nielder},
and \texttt{nzabikh}.
Each line in each of these files contains a single temporal interval as a text
string specifying a verb and two zero-origin nonnegative integers specifying
the starting and ending frames of the interval inclusive.
The LCA dataset also contains five object-class annotation files for each of
the video files in Table~\ref{tab:files}.
These also share the filename with the corresponding video, but with the
addition of the suffix \texttt{-enter-exits.txt}, and are located in the same
directories named with each of the above annotator codes.
Each line in each of these files contains a text string specifying an object
class, an integer specifying the number of objects of that class entering or
exiting the field of view (positive for entering and negative for exiting), and
a single zero-origin nonnegative integer specifying the video frame.

\section{Analysis}
\label{sec:analysis}

We analyzed the degree of agreement between the different annotators.
To do this, we compared pairs of annotators, taking the judgments of one as
`ground truth' and computing the F1 score of the other.
An interval in the annotation being scored was taken as a true positive if it
overlapped some interval with the same label in the `ground truth.'
An interval in the annotation being scored was taken as a false positive if it
didn't overlap any interval with the same label in the `ground truth.'
An interval in the `ground truth' was taken as a false negative if it didn't
overlap any interval with the same label in the annotation being scored.
From these counts, an F1 score could be computed.

\begin{figure*}
  \centering
  \includegraphics[width=0.9\textwidth]{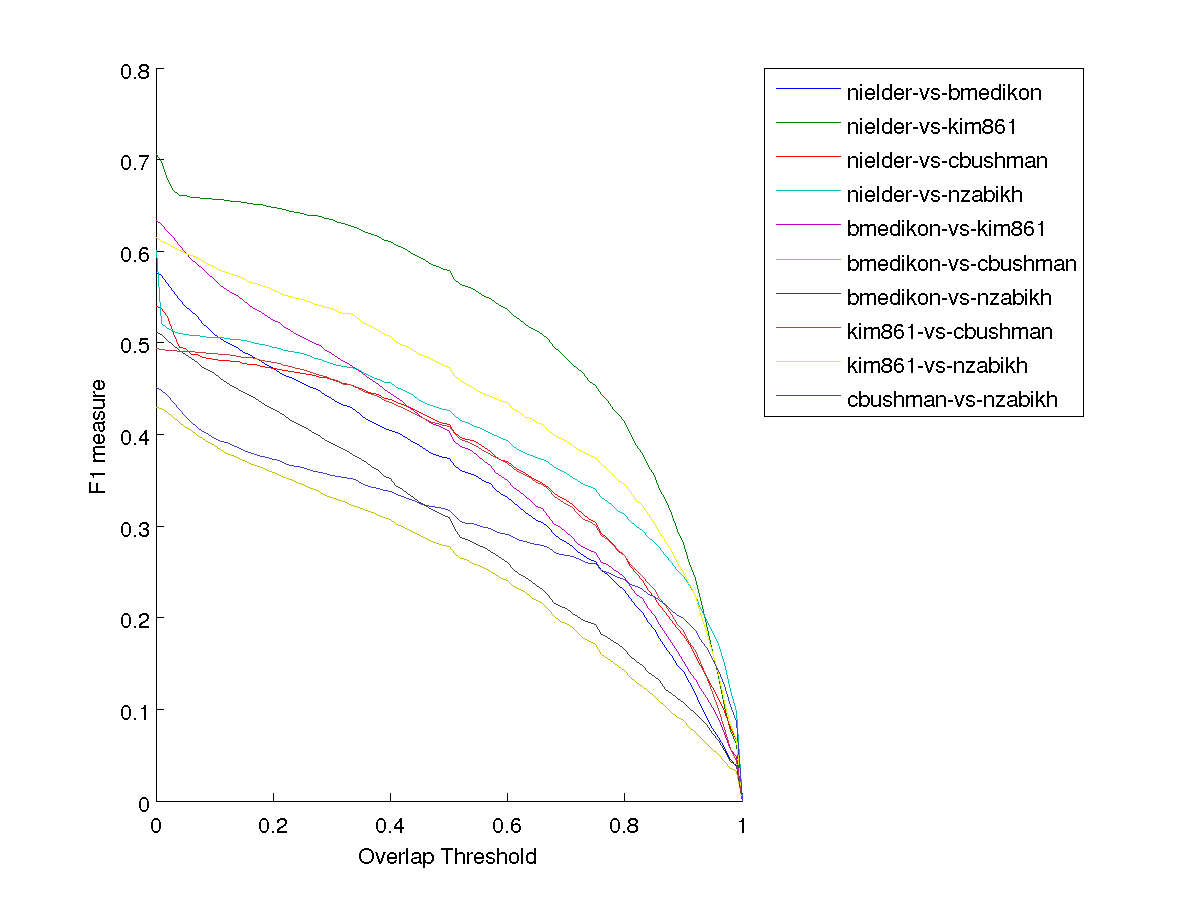}
  \caption{Intercoder agreement on the annotations of the LCA dataset.
    F1 score for each pair of annotators as the overlap criterion is varied.
    Overlap of two intervals is measured as the length of their intersection
    divided by the length of their union.}
  \label{fig:agreement}
\end{figure*}

We employed the following overlap criterion.
For a pair of intervals, we computed a one-dimensional variant of the
`intersection over union' criterion employed within the Pascal VOC Challenge
to determine overlap of two axis-aligned rectangles \citep{everingham10},
namely the length of the intersection divided by the length of the union.
We considered two intervals to overlap when the above exceeded some specified
threshold.
We then computed the F1 score as this threshold was varied and plotted the
results for all pairs of annotators (Fig.~\ref{fig:agreement}).

Note that there is a surprisingly low level of agreement between annotators.
Annotators rarely if ever agree on the precise temporal extent of an action as
indicated by the fact that all agreement curves go to zero as the overlap
threshold goes to one.
At an overlap threshold of 0.5, the F1 score varies between about 0.3 and about
0.6.
At an overlap threshold of 0.1, the threshold employed by VIRAT to score
machines against humans, the F1 score varies between about 0.38 and about 0.67.
This would put an upper bound on machine performance with this dataset using
the VIRAT threshold.
Even if the overlap threshold is reduced to zero, the F1 score varies between
about 0.43 and about 0.7.
This indicates that this dataset should be challenging for computer action
recognition.

This difficulty is corroborated by a recent paper \citep{barbu2014a}.
That paper employs a different subset of video from the DARPA Mind's Eye Year~2
evaluation that is extremely similar to that in the LCA dataset.
That dataset was annotated with the same procedures that were used to annotate
the LCA dataset.
The six verbs with highest intercoder agreement were selected: \emph{carry},
\emph{dig}, \emph{hold}, \emph{pick up}, \emph{put down}, and \emph{walk}.
For each of these, between 23 and 30 clips of 2.5s duration with the highest
level of intercoder agreement were selected, yielding 169 distinct clips.
Seven different state-of-the-art computer-vision action-recognition methods (C2
\citep{jhuang2007biologically}, Action Bank \citep{Sadanand2012}, Stacked ISA
\citep{le2011learning}, VHTK \citep{iccv2009MessingPalKautz}, Cao's
implementation \citep{cao2013recognizing} of Ryoo's method
\citep{ryoo2011human}, Cao's method \citep{cao2013recognizing}, and our own
implementation of the classifier described by Wang \etal\ \citep{dtijcv2013} on
top of the Dense Trajectories \citep{wang:2011:inria-00583818:1, dtijcv2013,
  Wang2013} feature extractor) were employed on this dataset, performing
one-of-out-six classification in an eight-fold cross-validation.
Note that for this task, each 2.5s clip was labeled with precisely one of the
six verbs as ground truth.
All seven methods performed extremely poorly on this dataset (C2 47.4\%,
Action Bank 44.2\%, Stacked ISA 46.8\%, VHTK 32.5\%, Cao's
implementation of Ryoo's method 31.2\%, Cao's method 33.3\%, and Dense
Trajectories 52.3\%), a task with only six classes and chance performance of
16.6\%.

\section{Baseline Experiment}

We performed an experiment similar to that of Barbu \etal\ \citep{barbu2014a} to
present and compare the performance of several state-of-the-art
action-recognition systems on the LCA dataset.
We evaluated all known action-recognition systems for which the code for the
end-to-end system is available, as well as implementations of several for which
the code is unavailable.
We used the same set of seven state-of-the-art action-recognition systems
compared in Barbu \etal\ \citep{barbu2014a} (C2 \citep{jhuang2007biologically},
Action Bank \citep{Sadanand2012}, Stacked ISA \citep{le2011learning}, VHTK
\citep{iccv2009MessingPalKautz}, Cao's implementation
\citep{cao2013recognizing} of Ryoo's method \citep{ryoo2011human}, Cao's method
\citep{cao2013recognizing}, and our own implementation of the classifier
described by Wang \etal\ \citep{dtijcv2013} on top of the Dense Trajectories
\citep{wang:2011:inria-00583818:1, dtijcv2013} feature extractor)
We also compared against our own implementation of the classifier described by
Wang \etal\ \citep{dtijcv2013} on top of the more recent Improved Trajectories
method \citep{Wang2013}.

As these methods are designed for classification of video clips, rather than
for streaming video, this experiment was performed on a subset of the LCA
dataset.
This subset was designed to be similar in character to other action-recognition
datasets and comprised short video clips.
It was created as follows.
First, we took the human-annotated action intervals produced by one of the
annotators, \texttt{cbushman}.
This annotator was chosen to maximize the number of available action intervals.
Next, a maximum of 100 intervals were selected for each action class.
For those action classes for which more than 100 intervals were annotated, a
random subset of 100 intervals was selected.
For those action classes with 100 or fewer annotated intervals, all annotated
intervals were used.
A 2s clip was extracted from the original videos centered in time on the
middle of each selected annotation interval.
These clips were temporally downsampled to 20 fps and spatially downsampled to a
width of 320 pixels, maintaining the aspect ratio.
This process resulted in a total of 1858 clips used for the baseline
experiment.

The class label of each clip was considered to be the action class
corresponding to the human-annotated interval from which the clip was derived.
The clips for each class were randomly split into a training set with 70\%
of the clips and a test set with 30\% of the clips, under the constraint
that sets of clips extracted from the same video should fall completely into
either the training or test set.
This was done to avoid having clips from the same action (\eg\ two clips from
the same person digging in the same location) from appearing in both the
training and test sets.
This resulted in a training set of 1318 training clips and 540 test clips.
Each method was trained on the training set and used to produce labels on the
test set.
All methods were run with default or recommended parameters.
These labels were compared to the intended class labels to measure the accuracy
of each method.
The results of this experiment are summarized in Table \ref{tab:results}.

\begin{table}
\caption{Comparison of accuracy for state-of-the-art action-recognition systems
  on a subset of the LCA dataset.}
\label{tab:results}
  \centering
  \resizebox{\columnwidth}{!}{\begin{tabular}{l|r}
      Method  & accuracy (\%)\\
      \hline
      Action Bank \citep{Sadanand2012} & 16.667 \\
      Improved Trajectories \citep{Wang2013} & 15.556 \\
      Dense Trajectories \citep{wang:2011:inria-00583818:1, dtijcv2013} &
       14.074 \\
      C2 \citep{jhuang2007biologically}  & 9.259\\
      Cao \citep{cao2013recognizing} & 7.592\\
      Cao's \citep{cao2013recognizing} implementation of Ryoo
      \citep{ryoo2011human} & 6.667\\
      Stacked ISA \citep{le2011learning} & 6.667\\
      VHTK \citep{iccv2009MessingPalKautz} & 6.296\\
      \hline
      chance (30/540) & 5.555\\
    \end{tabular}}
\end{table}

There are several things of note in these results.
First, all the accuracies are quite low, indicating the difficulty of the LCA
dataset.
The highest performing method, Action Bank, is correct only 16.667\% of the
time.
The four lowest performing methods have accuracies approaching chance
performance (5.555\%).
Additionally, the newer methods do not necessarily outperform the older
methods.
C2 significantly outperforms four more recently published methods, while Action
Bank is the best, outperforming even Improved Trajectories, which has the
highest performance on several well known datasets including HMDB (57.2\% vs
Action Bank's 26.9\%) and UCF50 (91.2\%).
We suspect that this difference in relative performance compared to other
datasets is the result of the lack of correlation between background and action
class which is often present in other datasets, as well as the presence of
multiple people in the field of view and the small relative size of the people
in the field of view.
That the performance is so low and that the highest scoring methods on other
datasets are not necessarily the same here shows that this dataset presents
new and difficult challenges not present in other datasets.

\section{Conclusion}
\label{sec:conclusion}

Upon acceptance of this manuscript we will make available to the community a
new dataset to support action-recognition research.
This dataset has more hours of video than HMDB51, roughly the same
amount of video as UCF50, about half as much video as UCF101 and Hollywood-2,
but unlike these has streaming video and has about twice as much video and
twice as many classes as VIRAT, the largest dataset of streaming video.
A distinguishing characteristic of this dataset is that the video is streaming;
long video segments contain many actions that start and stop at arbitrary
times, often overlapping in space and/or time.
A further distinguishing characteristic is that while all actions were filmed
in a variety of backgrounds, every action occurs in every background so that
background gives little information as to action class.
We employed novel techniques to annotate the temporal extent of action
occurrences.
A multiplicity of human annotations allows measuring intercoder agreement.
The above characteristics together with the surprisingly low level of
intercoder agreement suggest that this will be a challenging dataset.
This is confirmed by the low performance of recent methods on a baseline
experiment which also shows that those methods which perform best on other
datasets do not necessarily outperform other methods on this dataset.
The new difficulties posed by this dataset should spur significant advances in
action-recognition research.

\begin{acknowledgements}
This research was sponsored by the Army Research Laboratory and was
accomplished under Cooperative Agreement Number W911NF-10-2-0060.
The views and conclusions contained in this document are those of the authors
and should not be interpreted as representing the official policies, either
express or implied, of the Army Research Laboratory or the U.S. Government.
The U.S. Government is authorized to reproduce and distribute reprints for
Government purposes, notwithstanding any copyright notation herein.
\end{acknowledgements}

\bibliographystyle{spmpsci}
\bibliography{mva2014}

\end{document}